\documentclass[11pt]{article}

\usepackage[final]{acl}

\usepackage{kotex}
\usepackage{amsmath}
\usepackage{amssymb}
\usepackage{algorithm}
\usepackage{algorithmic}
\usepackage{adjustbox}

\usepackage{booktabs}
\usepackage{siunitx}
\usepackage{tabularx}
\usepackage{multirow}

\usepackage[table]{xcolor}
\usepackage{colortbl}

\usepackage{graphicx}
\usepackage{subcaption}

\usepackage{times}
\usepackage{latexsym}

\usepackage[T1]{fontenc}

\usepackage[utf8]{inputenc}

\usepackage{microtype}

\usepackage{inconsolata}

\usepackage{graphicx}

\title{TaRA: Training-Aware Low-Rank Adaptation Initialization}

\author{
 \textbf{Taehyeon Kim}\textsuperscript{\normalfont 1} \quad
 \textbf{Eunhyeok Park}\textsuperscript{\normalfont 2} \quad \\
 \textsuperscript{1}Department of Computer Science and Engineering \\
 \textsuperscript{2}Graduate School of Artificial Intelligence \\
 Pohang University of Science and Technology (POSTECH) \\
 {\fontsize{10pt}{12pt}\selectfont
 \texttt{\{taehyeonkim, eh.park\}@postech.ac.kr}}
}

\newcommand{\ours}{TaRA}
\newcommand{\edit}[1]{\textcolor{black}{#1}}
\newcommand{\red}[1]{\textcolor{red}{#1}}
\newcommand{\orange}[1]{\textcolor{orange}{#1}}

\begin{document}
\maketitle

\begin{abstract}
Low-Rank Adaptation (LoRA) has become a de facto standard for parameter-efficient fine-tuning (PEFT), yet its performance is highly sensitive to initialization due to the information bottleneck imposed by low-rank decomposition. Existing approaches attempt to construct high-quality LoRA initializations by exploiting principal components of pretrained weights, activations, or gradients. However, these methods do not directly account for the training dynamics of the full-rank model. In this paper, we propose Training-aware Low-Rank Adaptation Initialization (TaRA), a method that initializes LoRA such that the gradients induced by the low-rank factors closely approximate the gradient of the corresponding full-rank weight matrix. Derived from a mathematical formulation, TaRA improves gradient fidelity at the start of training while introducing negligible computational overhead. Across diverse and challenging fine-tuning tasks, TaRA consistently outperforms prior state-of-the-art methods, establishing a simple, robust, and scalable solution for effective LoRA initialization.
\end{abstract}

\section{Introduction}

Parameter-efficient fine-tuning (PEFT) has significantly reduced the cost of adapting large language models (LLMs), enabling their widespread deployment in practical applications \cite{houlsby2019parameter, li2021prefix}. Among PEFT methods, Low-Rank Adaptation (LoRA) \cite{hu2022lora} has emerged as the most widely adopted approach due to its simplicity, substantial reduction in fine-tuning resource requirements, and absence of inference-time overhead, as the learned updates can be merged into the base weights after training. These advantages have established LoRA as a de facto standard for efficient LLM fine-tuning.

\begin{figure}[t]
\centering
  \includegraphics[width=1\columnwidth]{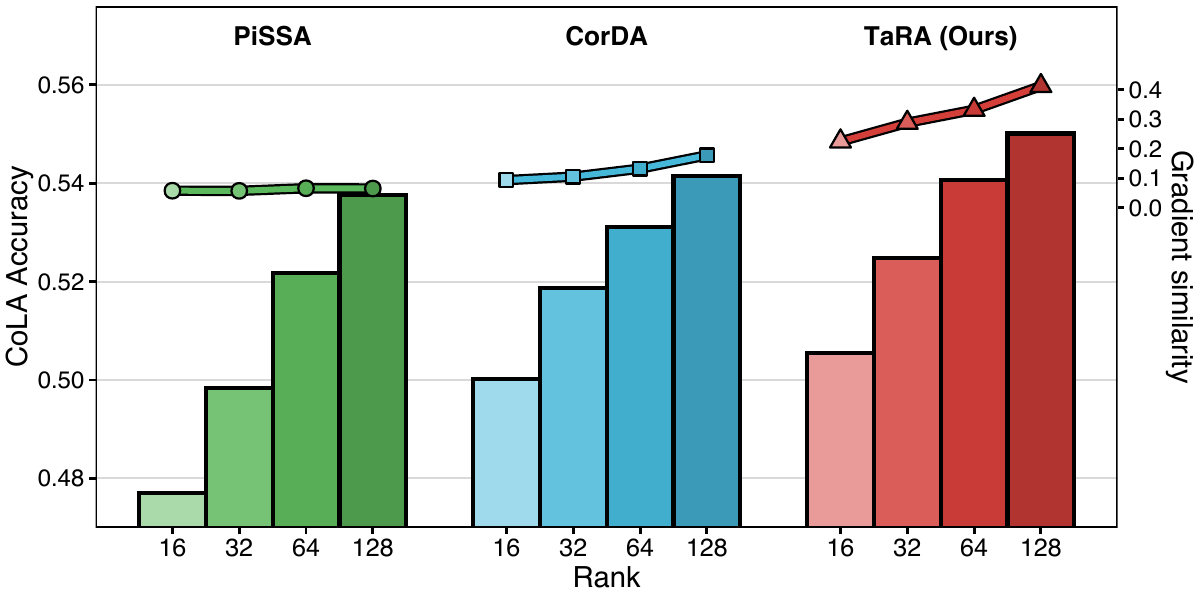}
  \caption{
Gradient similarity, measured as the cosine similarity between the one-step gradient of the combined LoRA adapter and that of the full weight, correlates with downstream accuracy.}

  \label{fig:motivation}
\end{figure}

Despite these benefits, LoRA introduces an inherent trade-off: the structured bottleneck imposed by low-rank decomposition alters the optimization trajectory relative to full-rank training, which can lead to suboptimal convergence even when model capacity is sufficient. Consequently, improving the effectiveness of LoRA without sacrificing its efficiency has become an important research direction \cite{meng2024pissa,buyukakyuz2406olora,hayou2024impactofinit}.

One promising direction is to design more informed initialization strategies for the low-rank adapters. While the original LoRA initializes its adapters with random or zero values, subsequent studies have shown that low-rank initializations guided by pretrained weight distributions (e.g., PiSSA \cite{meng2024pissa}), joint weight-data statistics (e.g., CorDA \cite{yang2024corda}), or gradients from a single forward-backward pass (e.g., LoRA-GA, LoRA-One \cite{wang2024loraga,zhang2025loraone}) consistently yield improved fine-tuning performance. These methods preserve the practical advantages of LoRA while alleviating degradation caused by low-rank constraints.

To further address this limitation, we propose Training-aware Low-Rank Adaptation (\textbf{\ours}), a new LoRA initialization method. Unlike prior approaches that approximate tensor statistics, \ours~is designed to preserve directions that are important to the local gradient behavior of full-weight training, allowing the low-rank parameterization to better reflect training-relevant information at initialization.
Grounded in a mathematical formulation, \ours~derives a low-rank initialization that preserves training-relevant directions in the local gradient field by jointly leveraging activation covariance, gradient covariance, and pretrained weights from a single forward-backward pass.
As illustrated in Figure \ref{fig:motivation}, this initialization achieves substantially higher gradient alignment than prior methods at the same rank, leading to consistent downstream improvements. Extensive experiments on challenging reasoning tasks, including mathematical problem solving, code generation, and commonsense reasoning, demonstrate that \ours~outperforms existing approaches.

\section{Prior Work on LoRA Initialization}

\begin{figure*}[t]
\centering
  \includegraphics[width=2\columnwidth]{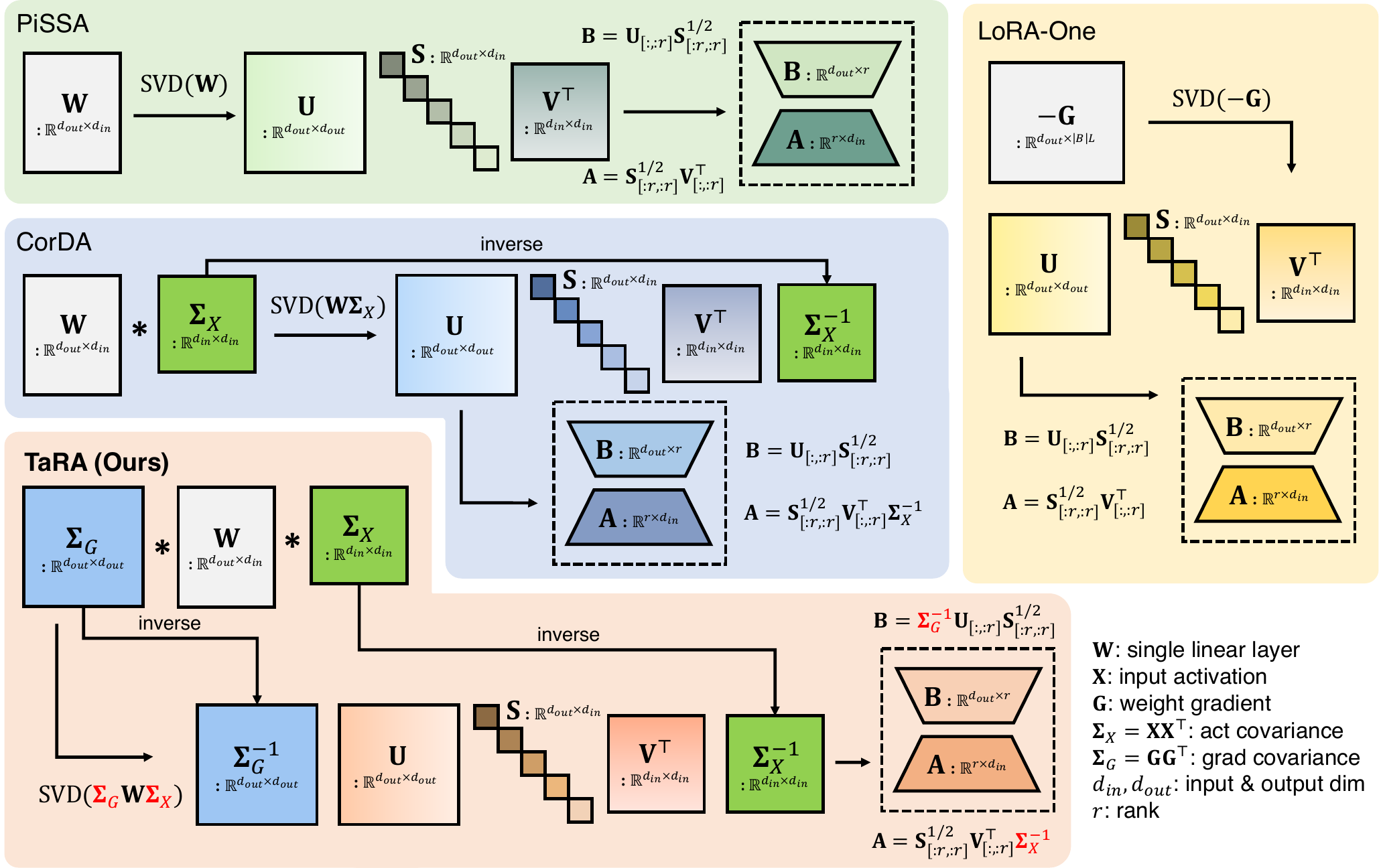}
\caption{
Illustration of our single-layer procedure.
We collect activations $\mathbf{X}$ and gradients $\mathbf{G}$ to form covariances $\mathbf{\Sigma}_X$ and $\mathbf{\Sigma}_G$.
PiSSA applies SVD to $\mathbf{W}$; CorDA applies SVD to $\mathbf{W}\mathbf{\Sigma}_X$ and maps back with $\mathbf{\Sigma}_X^{-1}$; LoRA-One applies SVD to $-\mathbf{G}$.
In contrast, \ours~ applies SVD to $\mathbf{\Sigma}_G\mathbf{W}\mathbf{\Sigma}_X$ and maps back with $\mathbf{\Sigma}_G^{-1}$ and $\mathbf{\Sigma}_X^{-1}$.
}
  \label{fig:main_figure}
\end{figure*}

In the original LoRA, the input-side factor matrix $\mathbf{A}$ is randomly initialized, while the output-side factor matrix $\mathbf{B}$ is set to zero. Although this design ensures that fine-tuning starts from the pretrained representations, the resulting initialization contains no meaningful task-related information, which can lead to slow early-stage optimization and suboptimal convergence \cite{wang2024loraga}.

Motivated by these limitations, numerous studies have explored improved initialization strategies for LoRA (Figure \ref{fig:main_figure}). Although the decomposed path produces a non-zero update, fine-tuning can still start from the pretrained weight $\mathbf{W}_{0}$ by absorbing this modification into a frozen residual weight, defined as $\mathbf{W}_{res} = \mathbf{W}_{0} - \mathbf{B}\mathbf{A}$. With an appropriate choice of $\mathbf{A}$ and $\mathbf{B}$, a properly initialized model can accelerate training and typically converges to better solutions than random initialization.

A key early work in this line of research is PiSSA \cite{meng2024pissa}, which proposes a data-agnostic initialization that leverages the pretrained weight matrix $\mathbf{W}_{0} \in \mathbb{R}^{d{out} \times d_{in}}$. Specifically, PiSSA performs singular value decomposition (SVD), $\mathbf{W}_{0} = \mathbf{U}\mathbf{S}\mathbf{V}^{\top}$, and initializes LoRA using the top-$r$ singular components:
\begin{align}
\mathbf{B} &= \mathbf{U}{[:,:r]}\mathbf{S}^{1/2}{[:r,:r]} \in \mathbb{R}^{d{out}\times r},\\
\mathbf{A} &= \mathbf{S}^{1/2}{[:r,:r]}\mathbf{V}^{\top}{[:,:r]} \in \mathbb{R}^{r\times d_{in}}.
\end{align}

More recently, CorDA \cite{yang2024corda} extends this idea by incorporating input statistics in addition to pretrained weights. CorDA first estimates the input activation covariance $\mathbf{\Sigma}_{X} = \mathbf{X}\mathbf{X}^{\top}$ from input activations $\mathbf{X}$, and then performs the following decomposition:
\begin{equation}
\text{SVD}(\mathbf{W}_{0}\mathbf{\Sigma}_{X})\mathbf{\Sigma}_{X}^{-1}
= \mathbf{\bar{U}}\mathbf{\bar{S}}\mathbf{\bar{V}}^{\top}.
\end{equation}
The resulting components are used to initialize $\mathbf{B}$, $\mathbf{A}$, and $\mathbf{W}_{res}$ in the same manner as PiSSA. Both PiSSA and CorDA align the LoRA subspace with principal components derived from pretrained weights or input-dependent statistics, resulting in faster convergence and improved accuracy.

Orthogonal to these approaches, LoRA-GA \cite{wang2024loraga} and its successor, LoRA-One \cite{zhang2025loraone}, leverage gradient information. These methods are motivated by the observation that the fine-tuning quality of LoRA depends on the alignment of its update direction with the dominant singular subspace of the full fine-tuning gradient. Based on this insight, LoRA-One performs SVD of the one-step full gradient of weight $\mathbf{G}$,
\begin{equation}
    \text{SVD}(-\mathbf{G})=\mathbf{\hat{U}}\mathbf{\hat{S}}\mathbf{\hat{V}}^{\top},
\end{equation}
and use the top-$r$ singular components to initialize the LoRA.
By leveraging gradient information, these methods initialize the LoRA modules along directions that are partially informed by training signals, often leading to improved fine-tuning performance in practice.

However, despite utilizing internal tensor statistics of the pretrained model, prior approaches primarily focus on enhancing the representational capacity of the low-rank decomposition, rather than explicitly aligning it with the learning dynamics of full fine-tuning. Consequently, a performance gap remains, which cannot be effectively closed by merely increasing the rank or tuning hyperparameters.

\section{Proposed Idea: TaRA}

In this section, we introduce TaRA, a novel LoRA initialization method designed to enhance fine-tuning quality under a strict low-rank constraint. We begin by formalizing the objective, derive a low-rank solution, and then describe its practical implementation.

\subsection{Motivation}

The key motivation behind TaRA is simple yet fundamental: even under the low-rank bottleneck imposed by LoRA, we aim to construct a low-rank parameterization whose induced local gradient behavior closely approximates that of the corresponding full-rank weight.
In other words,
TaRA embeds training-relevant directions of the local gradient field into the low-rank structure.
This perspective naturally leads to the following question:
\emph{\textbf{Under a rank-$r$ constraint, how can we construct a low-rank parameterization that preserves the directions most relevant to the local gradient behavior of full fine-tuning?}}

To formalize this intuition, we consider the following objective:
\begin{equation}
\begin{split}
  \tilde{\theta} = \arg\min_{\theta}
  \left\lVert\nabla\mathcal{L}(\theta)
  - \nabla\mathcal{L}(\theta_0)\right\rVert^2_F, \\
  \quad \text{s.t. } \mathrm{rank}(\theta) \leq r,
\end{split}
\label{eq:minimize_gradient_change}
\end{equation}

\noindent
where $\mathcal{L}$ denotes the task loss. Intuitively, this objective seeks a rank-constrained parameterization that preserves the local gradient behavior of the corresponding full-rank parameter at $\theta_0$.

\subsection{Training-Relevant Decomposition}
\label{sec:training-relevant-decomposition}

To obtain a tractable formulation, we adopt a standard second-order view of the loss landscape around the initialization point. In particular, we focus on the one-step gradient at initialization \cite{wang2024loraga}, which captures the dominant training signal in the early stage of optimization. In practice, this approximation is sufficient to guide effective low-rank initialization, as demonstrated by the convergence results in Section~\ref{sec:natural_language_generation_task}.

Using a second-order Taylor expansion of the loss around $\theta_0$, we obtain
\begin{equation}
\begin{split}
\mathcal{L}(\theta) \approx{} & \mathcal{L}(\theta_{0}) + \nabla\mathcal{L}(\theta_{0})^{\top}(\theta-\theta_{0}) \\
& + \frac{1}{2}(\theta-\theta_{0})^{\top}\mathbf{H}(\theta-\theta_{0}),
\end{split}
\label{eq:taylor_expansion}
\end{equation}

\noindent
where $\mathbf{H}$ denotes the Hessian evaluated at $\theta_{0}$.

In large models, directly computing the Hessian is infeasible. Following common practice, we instead use the Fisher information matrix $\mathcal{F}$ as a tractable surrogate for local curvature \cite{chekalina2025gfwsvd,martens2020fisherapprox}. Using the K-FAC factorization \cite{martens2015optimizing}, the Fisher matrix can be written as

\begin{equation}
  \mathcal{F} \approx \mathbf{\Sigma}_{X} \otimes \mathbf{\Sigma}_{G},
\end{equation}

\noindent
where $\mathbf{\Sigma}_{G} = \mathbf{G}\mathbf{G}^\top$ denotes the gradient covariance. Under this formulation, differentiating Eq.~(\ref{eq:taylor_expansion}) with respect to $\theta$ yields (derivation is in Appendix):

\begin{equation}
\begin{split}
\nabla\mathcal{L}(\theta) - \nabla\mathcal{L}(\theta_{0})
\approx
\mathbf{\Sigma}_{G}(\theta-\theta_{0})\mathbf{\Sigma}_{X}.
\end{split}
\label{eq:derivation_of_taylor_expansion}
\end{equation}

This relation reveals a key insight: the change in gradient induced by a parameter update is modulated by the curvature structure of the loss. Consequently, directions associated with large Fisher or Hessian values have a disproportionately large influence on the optimization trajectory. Preserving these directions is therefore crucial for faithfully approximating full fine-tuning under a low-rank space.

Empirically, the Hessian spectrum of modern neural networks is known to exhibit a highly structured form, consisting of a small set of prominent outlier eigenvectors and a near-zero bulk that is largely orthogonal to them \cite{gur2018gradient,sagun2017empirical}. Moreover, prior work shows that gradient descent effectively evolves within a small subspace spanned by the top Hessian eigenvectors \cite{gur2018gradient}. These observations suggest that dominant curvature directions provide a natural criterion for identifying effective low-rank training updates.

Motivated by this insight, we seek a rank-$r$ approximation $\tilde{\theta}$ that minimizes the curvature-weighted gradient deviation in Eq.~(\ref{eq:derivation_of_taylor_expansion}):

\begin{equation}
\begin{split}
\tilde{\theta} \approx \arg\min_{\theta}
\lVert\mathbf{\Sigma}_{G}(\theta-\theta_{0})\mathbf{\Sigma}_{X}\rVert^2_F,
\\ \quad \text{s.t. } \text{rank}(\theta) \leq r.
\end{split}
\label{eq:minimize_training_relevant_directions}
\end{equation}

This corresponds to a classical low-rank matrix approximation problem. By the Eckart--Young theorem \cite{eckart1936approximation,golub2013matrix}, the optimal solution is obtained by retaining the top-$r$ singular components of decomposition:

\begin{equation}
  \begin{split}
  \tilde{\theta} \approx \mathbf{\Sigma}_{G}^{-1}\operatorname{SVD}_r(\mathbf{\Sigma}_{G}\theta_{0}\mathbf{\Sigma}_{X})\mathbf{\Sigma}_{X}^{-1},
  \end{split}
  \label{eq:final_low_rank_approximation}
\end{equation}

\noindent
where $\operatorname{SVD}_{r}(\cdot)$ denotes the truncated SVD.\footnote{\edit{This is often confused with Fisher-weighted model compression; key differences are described in Appendix \ref{sec:difference_between_compression_and_ours}.}}

This result shows that the optimal low-rank approximation corresponds to preserving the dominant curvature directions of the pretrained parameter $\theta_0$.

\subsection{Implementation of \ours}

\begin{algorithm}[t]
  \caption{\ours, init for single layer}
  \label{alg:init_alg}
  \begin{algorithmic}
    \STATE \textbf{Input:} Pre-trained weight of single linear layer $\mathbf{W}_{0}\in\mathbb{R}^{d_{out}\times d_{in}}$, calibration dataset $D$, LoRA rank $r\in\mathbb{N}$
    \STATE \textbf{Collect:} \\
    \quad 1: Collect input activations in forward pass \\
    \quad \quad $\mathbf{X}\in\mathbb{R}^{d_{in}\times |B|L}$ \\
    \quad 2: Collect weight gradients in backward pass \\
    \quad \quad $\mathbf{G}\in\mathbb{R}^{d_{out}\times |B|L}$ \\
    \quad 3: Compute activation covariance \\
    \quad \quad $\mathbf{\Sigma}_{X}\leftarrow\mathbf{X}\mathbf{X}^{\top}\in\mathbb{R}^{d_{in}\times d_{in}}$ \\
    \quad 4: Compute gradient covariance \\
    \quad \quad $\mathbf{\Sigma}_{G}\leftarrow\mathbf{G}\mathbf{G}^{\top}\in\mathbb{R}^{d_{out}\times d_{out}}$
    \STATE \textbf{Init:} \\
    \quad 1: $\tilde{\mathbf{U}}, \tilde{\mathbf{S}}, \tilde{\mathbf{V}}^{\top} \leftarrow \text{SVD}(\mathbf{\Sigma}_{G}\mathbf{W}_{0}\mathbf{\Sigma}_{X})$ \\
    \quad 2: $\mathbf{B} \leftarrow \mathbf{\Sigma}_{G}^{-1}\tilde{\mathbf{U}}_{[:,:r]}\tilde{\mathbf{S}}^{1/2}_{[:r,:r]}$ $\in\mathbb{R}^{d_{out}\times r}$ \\
    \quad 3: $\mathbf{A} \leftarrow \tilde{\mathbf{S}}^{1/2}_{[:r,:r]}\tilde{\mathbf{V}}^{\top}_{[:,:r]}\mathbf{\Sigma}_{X}^{-1}$ $\in\mathbb{R}^{r\times d_{in}}$ \\
    \quad 4: $\mathbf{W}_{res} \leftarrow \mathbf{W}_{0}-\mathbf{B}\mathbf{A}$ $\in\mathbb{R}^{d_{out}\times d_{in}}$
  \end{algorithmic}
\end{algorithm}

Based on the above formulation, \ours~can be implemented as an efficient layer-wise LoRA initialization procedure for linear layers, summarized in Algorithm~\ref{alg:init_alg}. In the \textbf{Collect} stage, we gather task-dependent statistics using a small calibration dataset $D$. During the forward pass, we record input activations and concatenate them to form $\mathbf{X} \in \mathbb{R}^{d_{\text{in}} \times |B|L}$. During the backward pass, we collect the corresponding weight gradients to obtain $\mathbf{G} \in \mathbb{R}^{d_{\text{out}} \times |B|L}$. These statistics are used to estimate the covariance matrices $\mathbf{\Sigma}_{X}$ and $\mathbf{\Sigma}_{G}$. In the \textbf{Init} stage, we compute the SVD of $\mathbf{\Sigma}_{G}\mathbf{W}_{0}\mathbf{\Sigma}_{X}$ and project the resulting components back to the original space using $\mathbf{\Sigma}_{G}^{-1}$ and $\mathbf{\Sigma}_{X}^{-1}$. This produces singular components that capture the training-relevant directions under the Fisher-weighted metric.

In practice, the covariance matrices may be rank-deficient \cite{yankun2025svdq,zhao2024galore}, which makes direct inversion unstable. To ensure numerical stability, we apply diagonal damping,
$\mathbf{\Sigma}_{G} \leftarrow \mathbf{\Sigma}_{G} + c\beta\mathbf{I}$ and
$\mathbf{\Sigma}_{X} \leftarrow \mathbf{\Sigma}_{X} + c\beta\mathbf{I}$,
a well-known technique in second-order and Fisher-based optimization methods \cite{martens2015optimizing, ledoit2012nonlinear}. Following \cite{yang2024corda}, we set $\beta$ to the mean singular value of each covariance matrix and use $c=10^{-2}$ (see Appendix~\ref{sec:diagonal_damping_analysis} for further analysis).

After computing the decomposition, we truncate it to rank $r$ and construct the LoRA factors. The trainable matrices $\mathbf{A}$ and $\mathbf{B}$ are initialized from the top-$r$ singular components, scaled by the inverse covariance matrices to align them with the curvature-weighted subspace. We then update the frozen residual weight so that the model initially preserves the pretrained function. Applying this procedure independently to each linear layer yields a training-friendly LoRA initialization.

\newcommand{\std}[1]{\,{\textcolor{gray}{\scriptsize $\pm$ #1}}}

\begin{table*}[t]
\centering
\setlength{\tabcolsep}{5pt}
\renewcommand{\arraystretch}{1.12}
\scriptsize
\begin{tabular}{@{}clcccccc@{}}
\toprule
\textbf{Rank} & \textbf{Method} & \textbf{GSM8K-D} & \textbf{GSM8K-COT} & \textbf{MATH} & \textbf{HumanEval} & \textbf{MBPP} & \textbf{AVG} \\
\midrule

\multirow{7}{*}{128}
& Full fine-tuning (lr=4e-5) & 54.71\std{0.20} & 50.70\std{0.25} & 10.15\std{0.04} & 26.83\std{0.50} & 26.20\std{0.65} & 33.72\std{0.33} \\
\midrule
& LoRA (lr=4e-5)             & 44.83\std{0.38} & 37.23\std{0.55} & 6.12\std{0.04}  & 21.14\std{0.29} & 22.67\std{0.77} & 26.40\std{0.41} \\
& PiSSA (lr=4e-5)            & 53.85\std{0.07} & 47.42\std{0.14} & 8.98\std{0.04}  & 23.46\std{0.64} & 24.87\std{0.57} & 31.72\std{0.29} \\
& CorDA (lr=4e-5)            & 55.54\std{0.54} & 47.56\std{0.27} & 9.41\std{0.11}  & \red{\textbf{24.06}}\std{0.20} & 24.10\std{0.22} & 32.13\std{0.27} \\
& LoRA-One (lr=2e-4)         & 52.92\std{0.49} & 46.74\std{0.22} & 9.28\std{0.07}  & 23.81\std{0.59} & 24.83\std{0.47} & 31.52\std{0.37} \\
& \textbf{\ours} (lr=4e-5)             & \red{\textbf{56.59}}\std{0.25} & \red{\textbf{50.42}\std{0.13}} & \red{\textbf{10.08}}\std{0.10} & 22.59\std{0.58} & \red{\textbf{25.20}}\std{0.67} & \red{\textbf{32.98}}\std{0.34} \\
\midrule

\multirow{5}{*}{64}
& LoRA (lr=4e-5)             & 40.31\std{0.18} & 32.65\std{0.20} & 5.27\std{0.05}  & 17.88\std{0.60} & 21.40\std{0.59} & 23.50\std{0.32} \\
& PiSSA (lr=4e-5)            & 49.05\std{0.86} & 43.57\std{0.28} & 7.29\std{0.16}  & 21.47\std{0.97} & 23.73\std{0.62} & 29.02\std{0.58} \\
& CorDA (lr=4e-5)            & 51.45\std{0.40} & 42.77\std{0.33} & 8.13\std{0.18}  & 21.14\std{0.76} & 24.93\std{0.41} & 29.68\std{0.42} \\
& LoRA-One (lr=2e-4)         & 51.54\std{0.83} & 42.84\std{0.43} & 8.06\std{0.08}  & \red{\textbf{24.39}}\std{0.86} & 25.20\std{0.43} & 30.41\std{0.53} \\
& \textbf{\ours} (lr=4e-5)             & \red{\textbf{54.12}}\std{0.93} & \red{\textbf{44.49}}\std{0.11} & \red{\textbf{9.23}}\std{0.07} & 23.58\std{0.70} & \red{\textbf{25.40}}\std{0.36} & \red{\textbf{31.36}}\std{0.43} \\
\midrule

\multirow{5}{*}{32}
& LoRA (lr=4e-5)             & 38.36\std{0.35} & 31.51\std{0.22} & 5.10\std{0.12}  & 18.90\std{0.68} & 20.47\std{0.82} & 22.87\std{0.44} \\
& PiSSA (lr=4e-5)            & 46.40\std{0.06} & 36.64\std{0.29} & 6.42\std{0.06}  & 20.73\std{0.50} & 22.31\std{0.13} & 26.50\std{0.21} \\
& CorDA (lr=4e-5)            & 48.09\std{0.22} & 38.82\std{0.64} & 6.96\std{0.06}  & 19.10\std{0.58} & 24.33\std{0.82} & 27.46\std{0.46} \\
& LoRA-One (lr=2e-4)         & 47.44\std{0.30} & 41.55\std{0.17} & 7.49\std{0.11}  & 20.69\std{0.26} & 24.60\std{0.43} & 28.35\std{0.26} \\
& \textbf{\ours} (lr=4e-5)             & \red{\textbf{50.14}}\std{0.08} & \red{\textbf{41.70}}\std{0.27} & \red{\textbf{8.57}}\std{0.06} & \red{\textbf{21.53}}\std{0.29} & \red{\textbf{24.73}}\std{0.52} & \red{\textbf{29.33}}\std{0.24} \\
\bottomrule
\end{tabular}
\caption{Comparison of LoRA, PiSSA, CorDA, LoRA-One, and \ours~on natural language generation tasks. Each result is the mean over three seeds, and the standard deviation is shown in small gray text. \red{Red} color indicates the best PEFT result within each model block, excluding Full FT.}
\label{tab:natural_language_generation_task_results}
\end{table*}

\section{Experiments}
\label{sec:experiemnt}

We evaluate the proposed method through extensive experiments across multiple LLMs and a diverse set of downstream tasks. We compare \ours~against full fine-tuning and several LoRA initialization baselines, including LoRA \cite{hu2022lora}, PiSSA \cite{meng2024pissa}, CorDA \cite{yang2024corda}, and LoRA-One \cite{zhang2025loraone}. We also include the state-of-the-art LoRA variants, MiSS \cite{kang2409miss} and LoRAM \cite{zhang2026loram} for comparison. Our evaluation covers two task categories: natural language generation (NLG) and natural language understanding (NLU), and systematically examines performance across different model and LoRA rank budgets.

\subsection{Experimental Setup}

\paragraph{Natural Language Generation Tasks.}

For NLG, we consider math problem solving and code generation. We fine-tune LLaMA-2-7B \cite{touvron2023llama2} on 100K samples from MetaMathQA \cite{yu2023metamath} for math and 100K samples from CodeFeedback-Filtered-Instruction \cite{zheng2024codefeedback} for code. Models are trained with LoRA ranks of 128, 64 and 32 to assess robustness across rank budgets (LoRA $\alpha$ is set equal to the rank).

For evaluation, math performance is measured on GSM8K-D \cite{cobbe2021gsm8k} and MATH \cite{hendrycks2021hendrycksmath} using direct prompting, as well as GSM8K-COT \cite{wei2022gsm8kcot} under an 8-shot chain-of-thought setting with the Language Model Evaluation Harness \cite{biderman2024lmevalharness}. Code generation is evaluated on HumanEval \cite{chen2021humaneval} and MBPP \cite{austin2021mbpp} using the BigCode Evaluation Harness \cite{bigcode-evaluation-harness}.
All evaluation harnesses are used with their default configurations.

\begin{figure*}[h]
  \centering
  \begin{subfigure}[t]{\columnwidth}
      \centering
      \includegraphics[width=\linewidth]{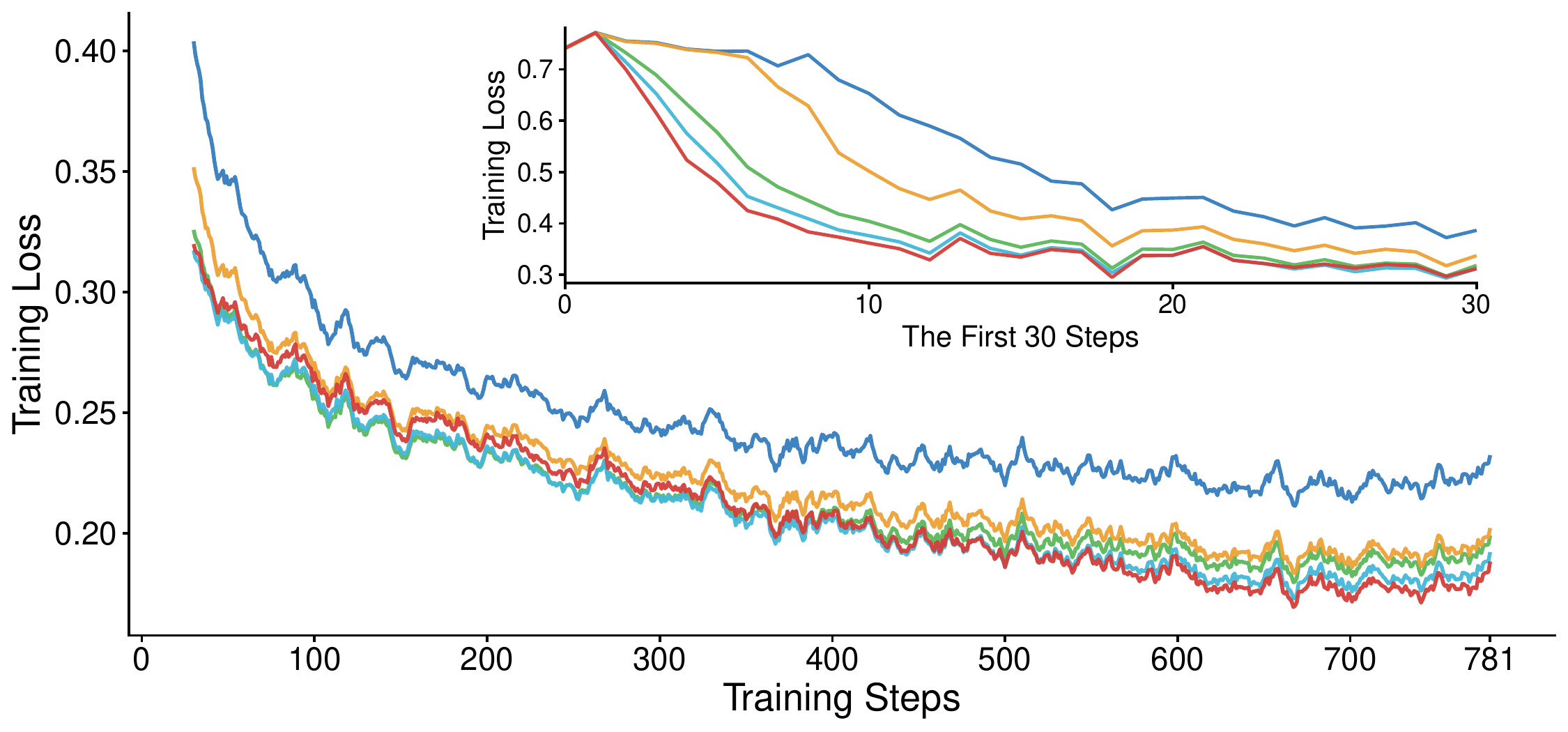}
      \caption{$r=128$}
      \label{fig:training_loss_rank128_math}
  \end{subfigure}
\begin{subfigure}[t]{\columnwidth}
  \centering
  \includegraphics[width=\linewidth]{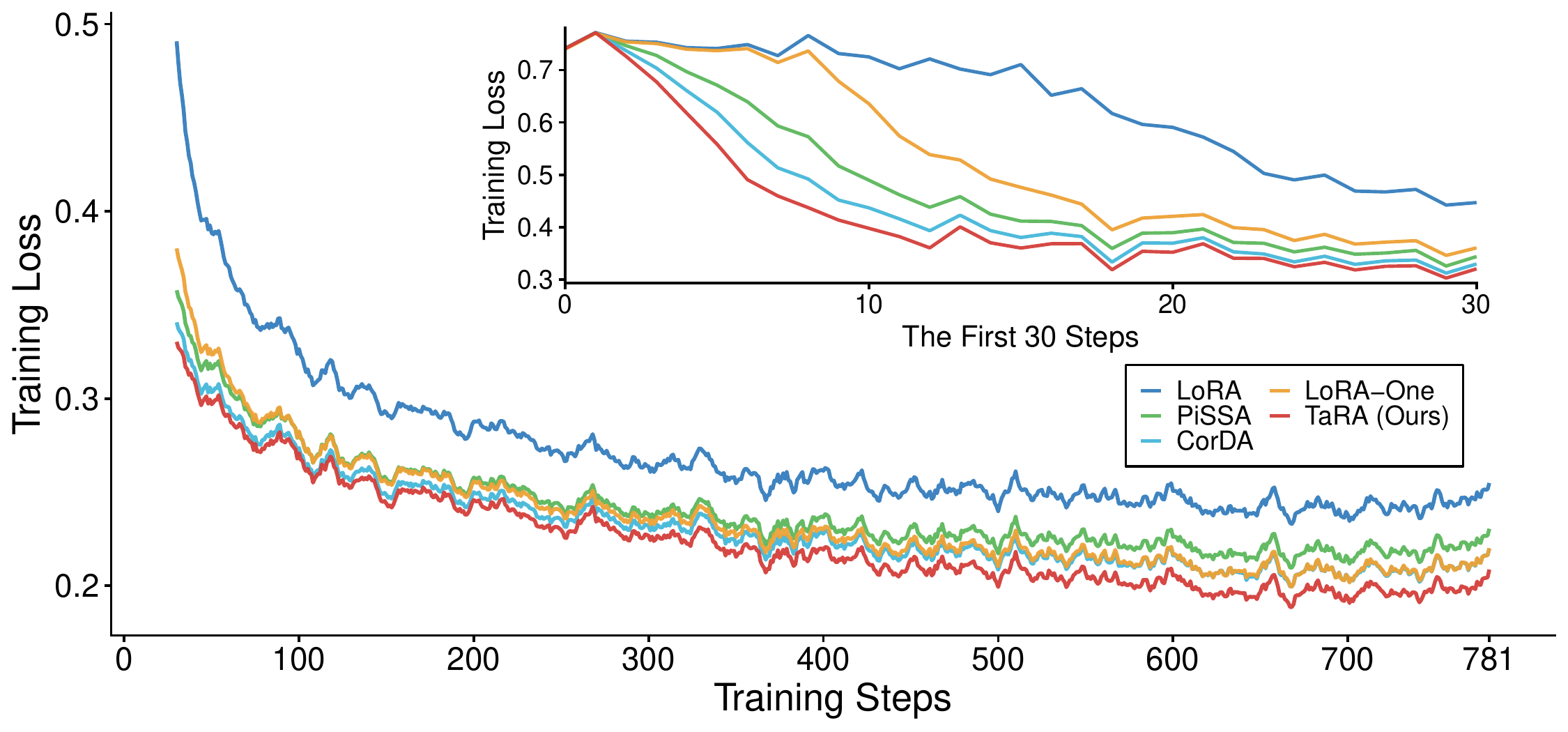}
  \caption{$r=32$}
  \label{fig:trianing_loss_rank32_math}
\end{subfigure}
\caption{Training loss over steps on LLaMA-2-7B fine-tuned with MetaMathQA. We compare LoRA, PiSSA, CorDA, LoRA-One, and \ours~at (a) $r=128$ and (b) $r=32$; insets show the first 30 steps.}
  \label{fig:training_loss_math}
\end{figure*}

\begin{figure}[t]
  \centering
    \includegraphics[width=\columnwidth]{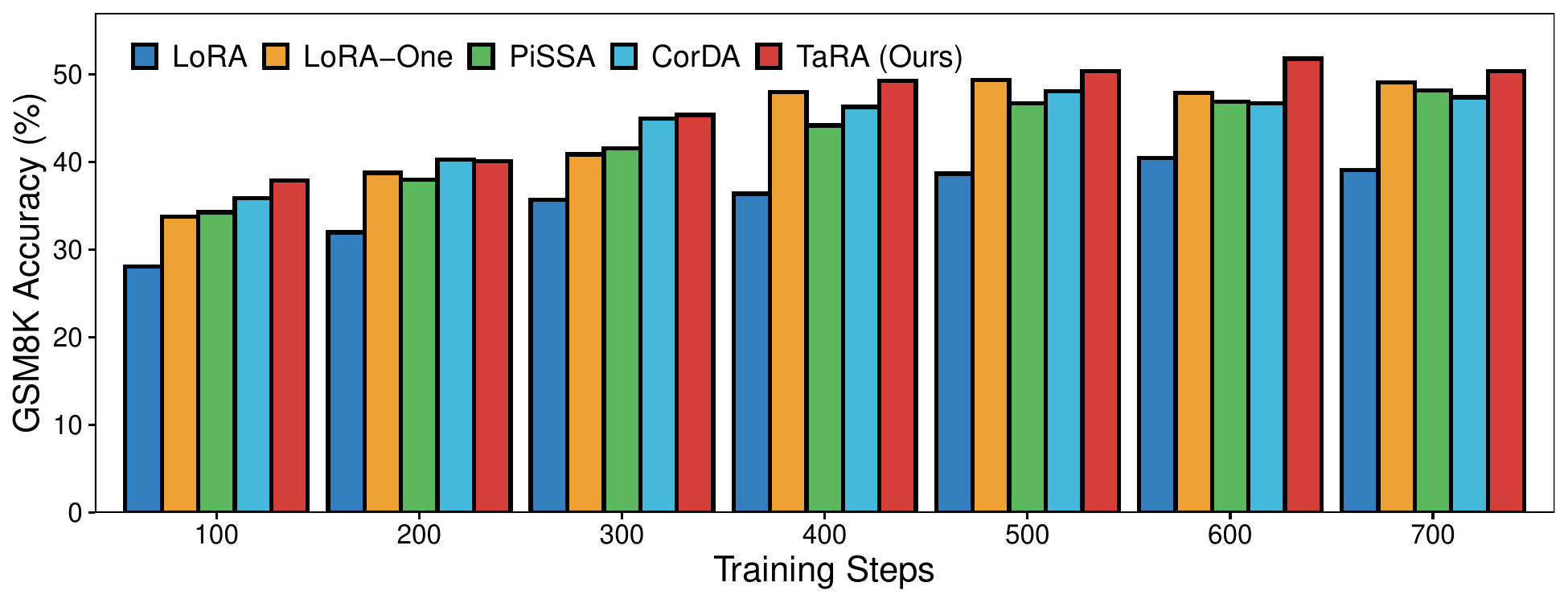}
  \caption{GSM8K-D accuracy over training steps when fine-tuning LLaMA-2-7B on the MetaMathQA dataset.}
    \label{fig:gsm8k_over_steps}
  \end{figure}

\paragraph{Natural Language Understanding Tasks.}

For NLU, we focus on commonsense reasoning. Models are fine-tuned on the Commonsense-170K dataset \cite{hu2023llm-adapters}. To evaluate scalability and generality, we conduct experiments on three base models—DeepSeek-R1-Distill-Qwen-1.5B \cite{guo2025deepseek-r1}, LLaMA-2-7b \cite{touvron2023llama2}, LLaMA-3.1-8B \cite{grattafiori2024llama3}, and Qwen-3-8B \cite{yang2025qwen3}—with a fixed LoRA rank of 128 (LoRA $\alpha$ is set equal to the rank).

We evaluate performance on eight benchmarks: BoolQ \cite{clark2019boolq}, PIQA \cite{bisk2020piqa}, SIQA \cite{sap2019socialiqa}, HellaSwag \cite{zellers2019hellaswag}, WinoGrande \cite{sakaguchi2021winogrande}, ARC-Challenge, ARC-Easy \cite{clark2018arc}, and OBQA \cite{mihaylov2018obqa}.

For full fine-tuning, LoRA, PiSSA, CorDA and \ours, we follow the hyperparameter settings of CorDA \cite{yang2024corda} and select the best learning rate. For LoRA-One, which is sensitive to learning-rate choice, we use the optimal value reported in its original paper. For all methods that require calibration (CorDA, LoRA-One, and \ours), we set the calibration set size to 256 and conducted all experiments under this setting. All results are obtained from the final training checkpoint and reported in terms of task accuracy. Additional details are provided in the Appendix. All experiments are conducted on A100 GPUs (80GB).

\subsection{Natural Language Generation Task}
\label{sec:natural_language_generation_task}

Table~\ref{tab:natural_language_generation_task_results} compares performance on natural language generation benchmarks across different LoRA ranks ($r \in {128, 64, 32}$).
Overall, \ours~achieves the highest average performance at all ranks, consistently outperforming prior initialization methods such as PiSSA, CorDA, and LoRA-One.
At $r=128$, \ours~attains the best results on GSM8K-D/GSM8K-COT, MATH, and MBPP, and also yields the highest average score.
At $r=64$, \ours~remains the top-performing method on GSM8K-D/GSM8K-COT, MATH, and MBPP, leading to the best average performance as well.
Even at $r=32$, \ours~exhibits a relatively small performance drop compared to other baselines, achieving the best performance across all tasks and demonstrating the most stable behavior across ranks.
These results highlight the strength of the proposed initialization in efficiently capturing training-relevant directions under a constrained parameter budget.

Notably, LoRA-One also underperforms PiSSA and CorDA at $r=128$, suggesting that directly applying SVD to gradients is ineffective at higher ranks. Overall, these results indicate that neither simple SVD-based nor random initialization is sufficient to capture training-relevant directions under tight constraints, whereas \ours~does so effectively.

Figures~\ref{fig:training_loss_rank128_math} and \ref{fig:trianing_loss_rank32_math} show the training loss curves on MetaMathQA for $r=128$ and $r=32$. In both settings, \ours~converges more smoothly and reaches a lower loss than prior methods.
This suggests that the proposed initialization provides a more effective update subspace at the early stage of training, enabling more efficient optimization, and that its benefit persists into later phases of training \cite{gur2018gradient}.
This advantage persists throughout fine-tuning, as also reflected in GSM8K-D accuracy over training steps (Figure~\ref{fig:gsm8k_over_steps}). While PiSSA and CorDA exhibit limited improvement and LoRA-One improves only in later stages, \ours~achieves rapid early gains and consistently attains the best performance at both intermediate and final stages.

\begin{table}[t]
\centering
\setlength{\tabcolsep}{3pt}
\renewcommand{\arraystretch}{1.12}
\scriptsize
\begin{tabular}{@{}lcccc@{}}
\toprule
\textbf{Method} & \textbf{Rank} & \textbf{Trainable param.} & \textbf{GSM8K-D} & \textbf{GSM8K-COT} \\
\midrule
LoRAM       & 128 & 319.8M (4.53\%) & 53.65\std{0.05} & 47.46\std{0.05} \\
MiSS        & 256 & 348.1M (4.91\%) & 54.13\std{0.03} & 48.67\std{0.02} \\
\textbf{\ours} & 128 & 319.8M (4.53\%) & \red{\textbf{56.59}}\std{0.25} & \red{\textbf{50.42}}\std{0.13} \\
\bottomrule
\end{tabular}
\caption{Comparison with LoRA variants on GSM8K-D and GSM8K-CoT. Baseline results report the mean and variance over three seeds. \red{Red} color indicates the best result in each task.}
\label{tab:recent_lora_variant_results}
\end{table}

\begin{table*}[t]
  \centering
  \footnotesize
  \setlength{\tabcolsep}{4.2pt}
  \renewcommand{\arraystretch}{1.05}
  \begin{adjustbox}{width=\textwidth}
  \begin{tabular}{@{}l l c c c c c c c c c@{}}
  \toprule
  \textbf{Model} & \textbf{Method} &
  \textbf{BoolQ} & \textbf{PIQA} & \textbf{SIQA} & \textbf{HS} &
  \textbf{WG} & \textbf{ARC-c} & \textbf{ARC-e} & \textbf{OBQA} & \textbf{AVG} \\
  \midrule

  \multirow{6}{*}{\shortstack[l]{DeepSeek-R1\\-Distill-Qwen\\-1.5B}}
  & Full FT
    & 63.04\std{0.12} & 61.41\std{0.38} & 52.58\std{0.02} & 27.40\std{0.10}
    & 49.41\std{0.49} & 45.22\std{0.36} & 61.38\std{0.16} & 42.47\std{0.62} & 50.36\std{0.18} \\
  & LoRA
    & 62.60\std{0.04} & 61.64\std{0.38} & 51.66\std{0.16} & 25.77\std{0.24}
    & 50.57\std{0.44} & 46.33\std{0.30} & 62.16\std{0.24} & 42.73\std{0.52} & 50.43\std{0.05} \\
  & PiSSA
    & 64.68\std{0.11} & 65.02\std{0.09} & 63.15\std{0.27} & \red{\textbf{38.84}}\std{0.31}
    & 57.75\std{0.16} & 54.38\std{0.42} & \orange{69.32}\std{0.03} & 54.60\std{0.16} & 58.47\std{0.03} \\
  & CorDA
    & \red{\textbf{64.89}}\std{0.13} & \orange{66.14}\std{0.11} & \orange{65.51}\std{0.11} & 37.04\std{0.28}
    & 58.27\std{0.23} & \red{\textbf{55.26}}\std{0.15} & 68.62\std{0.19} & \orange{56.20}\std{0.49} & \orange{58.99}\std{0.09} \\
  & LoRA-One
    & 63.77\std{0.18} & 64.33\std{0.27} & 65.01\std{0.34} & 31.33\std{0.14}
    & \orange{58.82}\std{0.29} & 51.91\std{0.38} & \red{\textbf{69.46}}\std{0.14} & 55.13\std{0.41} & 57.47\std{0.12} \\
  & \textbf{\ours}
    & \orange{64.77}\std{0.20} & \red{\textbf{67.28}}\std{0.88} & \red{\textbf{65.52}}\std{0.11} & \orange{38.18}\std{0.16}
    & \red{\textbf{60.22}}\std{0.70} & \orange{54.52}\std{0.43} & 68.43\std{0.16} & \red{\textbf{59.20}}\std{0.49} & \red{\textbf{59.77}}\std{0.36} \\
  \midrule

  \multirow{6}{*}{LLaMA-2-7B}
  & Full FT
    & 70.62\std{0.29} & 81.18\std{0.09} & 78.04\std{0.15} & 76.04\std{0.09}
    & 79.03\std{0.18} & 66.61\std{0.04} & 83.71\std{0.06} & 77.80\std{0.00} & 76.63\std{0.06} \\
  & LoRA
    & 63.72\std{0.21} & 77.24\std{0.09} & 72.66\std{0.16} & 56.50\std{0.10}
    & 69.09\std{0.16} & 61.66\std{0.10} & 79.81\std{0.16} & 64.33\std{0.34} & 68.13\std{0.01} \\
  & PiSSA
    & 70.34\std{0.15} & \orange{80.92}\std{0.14} & 78.22\std{0.05} & 85.83\std{0.01}
    & 78.61\std{0.07} & \red{\textbf{67.75}}\std{0.07} & \red{\textbf{83.67}}\std{0.06} & 77.73\std{0.09} & 77.89\std{0.01} \\
  & CorDA
    & 70.42\std{0.24} & \red{\textbf{80.94}}\std{0.20} & \orange{78.93}\std{0.09} & 83.43\std{0.12}
    & 80.42\std{0.34} & 66.95\std{0.20} & 83.16\std{0.07} & 77.60\std{0.28} & 77.73\std{0.02} \\
  & LoRA-One
    & \orange{71.22}\std{0.08} & 80.88\std{0.08} & 78.48\std{0.02} & \red{\textbf{89.19}}\std{0.00}
    & \orange{80.56}\std{0.07} & \orange{67.12}\std{0.04} & 82.91\std{0.04} & \orange{79.40}\std{0.00} & \orange{78.72}\std{0.02} \\
  & \textbf{\ours}
    & \red{\textbf{71.34}}\std{0.04} & 80.49\std{0.09} & \red{\textbf{79.00}}\std{0.19} & \orange{88.84}\std{0.03}
    & \red{\textbf{80.74}}\std{0.11} & 66.98\std{0.14} & \orange{83.23}\std{0.08} & \red{\textbf{80.60}}\std{0.28} & \red{\textbf{78.90}}\std{0.04} \\
  \midrule

  \multirow{6}{*}{LLaMA-3.1-8B}
  & Full FT
    & 74.39\std{0.15} & 88.45\std{0.07} & 80.35\std{0.04} & 95.52\std{0.00}
    & 85.98\std{0.08} & 82.17\std{0.07} & 92.31\std{0.02} & 84.60\std{0.16} & 85.47\std{0.03} \\
  & LoRA
    & 71.93\std{0.15} & 85.91\std{0.04} & 77.43\std{0.07} & 92.63\std{0.08}
    & 81.06\std{0.11} & 78.24\std{0.07} & 90.57\std{0.06} & 79.47\std{0.09} & 82.15\std{0.02} \\
  & PiSSA
    & 74.30\std{0.23} & 88.21\std{0.25} & 79.97\std{0.08} & 95.19\std{0.09}
    & 85.74\std{0.30} & 80.03\std{0.19} & 91.38\std{0.44} & 86.13\std{0.25} & 85.13\std{0.02} \\
  & CorDA
    & \orange{74.56}\std{0.23} & \orange{88.34}\std{0.23} & 80.16\std{0.26} & 95.21\std{0.02}
    & 85.37\std{0.24} & \red{\textbf{80.81}}\std{0.38} & \orange{91.74}\std{0.07} & \red{\textbf{86.60}}\std{0.16} & 85.35\std{0.06} \\
  & LoRA-One
    & 74.50\std{0.05} & 88.03\std{0.02} & \red{\textbf{80.34}}\std{0.24} & \red{\textbf{96.04}}\std{0.01}
    & \orange{86.35}\std{0.43} & \orange{80.66}\std{0.15} & 91.64\std{0.07} & 86.33\std{0.09} & \orange{85.49}\std{0.06} \\
  & \textbf{\ours}
    & \red{\textbf{74.69}}\std{0.27} & \red{\textbf{88.73}}\std{0.06} & \orange{80.33}\std{0.17} & \orange{95.35}\std{0.05}
    & \red{\textbf{86.72}}\std{0.32} & 80.49\std{0.29} & \red{\textbf{91.86}}\std{0.03} & \orange{86.40}\std{0.16} & \red{\textbf{85.57}}\std{0.04} \\
  \midrule

  \multirow{6}{*}{Qwen-3-8B}
  & Full FT
    & 72.07\std{0.09} & 89.74\std{0.09} & 80.19\std{0.13} & 93.19\std{0.02}
    & 81.09\std{0.20} & 91.86\std{0.11} & 96.59\std{0.06} & 88.87\std{0.50} & 86.70\std{0.06} \\
  & LoRA
    & \orange{71.39}\std{0.16} & 88.68\std{0.13} & 79.60\std{0.02} & 91.77\std{0.01}
    & 77.66\std{0.23} & \red{\textbf{91.53}}\std{0.08} & 96.56\std{0.02} & 88.13\std{0.09} & 85.67\std{0.04} \\
  & PiSSA
    & 71.34\std{0.30} & \orange{89.88}\std{0.12} & 80.09\std{0.18} & 94.25\std{0.01}
    & 82.95\std{0.13} & \orange{91.44}\std{0.04} & 96.86\std{0.04} & \orange{91.07}\std{0.19} & 87.24\std{0.03} \\
  & CorDA
    & 71.19\std{0.08} & 89.46\std{0.10} & \orange{81.19}\std{0.05} & 94.10\std{0.01}
    & 82.35\std{0.04} & 88.88\std{0.08} & 95.62\std{0.00} & 90.80\std{0.00} & 86.70\std{0.01} \\
  & LoRA-One
    & \red{\textbf{73.26}}\std{0.09} & 89.41\std{0.14} & 80.16\std{0.14} & \red{\textbf{94.57}}\std{0.01}
    & \red{\textbf{84.95}}\std{0.10} & 91.13\std{0.07} & \red{\textbf{97.22}}\std{0.00} & 90.40\std{0.28} & \red{\textbf{87.64}}\std{0.05} \\
  & \textbf{\ours}
    & 71.19\std{0.17} & \red{\textbf{89.90}}\std{0.07} & \red{\textbf{81.32}}\std{0.04} & \orange{94.38}\std{0.03}
    & \orange{83.11}\std{0.07} & 89.85\std{0.19} & \orange{96.88}\std{0.09} & \red{\textbf{91.40}}\std{0.00} & \orange{87.25}\std{0.02} \\
  \bottomrule
  \end{tabular}
  \end{adjustbox}
  \caption{Comparison of Full FT, LoRA, PiSSA, CorDA, LoRA-One, and \ours\ on commonsense reasoning tasks. Results are averaged over three seeds, with standard deviations shown in gray text. \red{Red} and \orange{Orange} color denote the best and second-best PEFT results in each model, excluding Full FT. Abbr. HS = HellaSwag, WG = WinoGrande.}
  \label{tab:commonsense_reasoning_benchmarks}
\end{table*}

\paragraph{Comparison with Non-Initialization PEFT variants}

To compare \ours~with recent PEFT variants beyond LoRA initialization methods, we evaluated MiSS and LoRAM on mathematical reasoning tasks. For both MiSS and LoRAM, we used the same learning rate as \ours, 4e-5. To approximately match the number of trainable parameters, we set the rank to 256 for MiSS and 128 for LoRAM. Table~\ref{tab:recent_lora_variant_results} presents the results. Overall, \ours~demonstrates that, even when using the standard LoRA parameterization, a well-designed initialization can outperform recent PEFT variants on mathematical reasoning tasks.

\subsection{Natural Language Understanding Task}

Table~\ref{tab:commonsense_reasoning_benchmarks} compares performance on standard commonsense reasoning benchmarks across three models: DeepSeek-R1-Distill-Qwen-1.5B, LLaMA-2-7B LLaMA-3.1-8B, and Qwen-3-8B. Overall, \ours~achieves the best average performance on DeepSeek-R1-Distill-Qwen-1.5B, LLaMA-2-7B and LLaMA-3.1-8B, outperforming PiSSA, CorDA, and LoRA-One. On Qwen-3-8B, \ours~slightly underperforms LoRA-One in terms of average score but remains competitive, consistently outperforming LoRA, PiSSA, and CorDA. While fine-tuning performance is often distribution-dependent and no method consistently dominates across all settings, a well-known phenomenon also observed in prior work, we evaluate methods based on their generality across diverse scenarios. Under this criterion, \ours~consistently performs well, achieving top-1 performance in 15 out of 32 cases and second-best in 11 more. In contrast, LoRA-One and CorDA achieve top-1 in only 8 and 5 cases, respectively, highlighting the robustness of \ours~beyond specific models or tasks.

\subsection{Gradient Alignment Analysis}

\begin{figure}[t]
  \centering
    \includegraphics[width=\columnwidth]{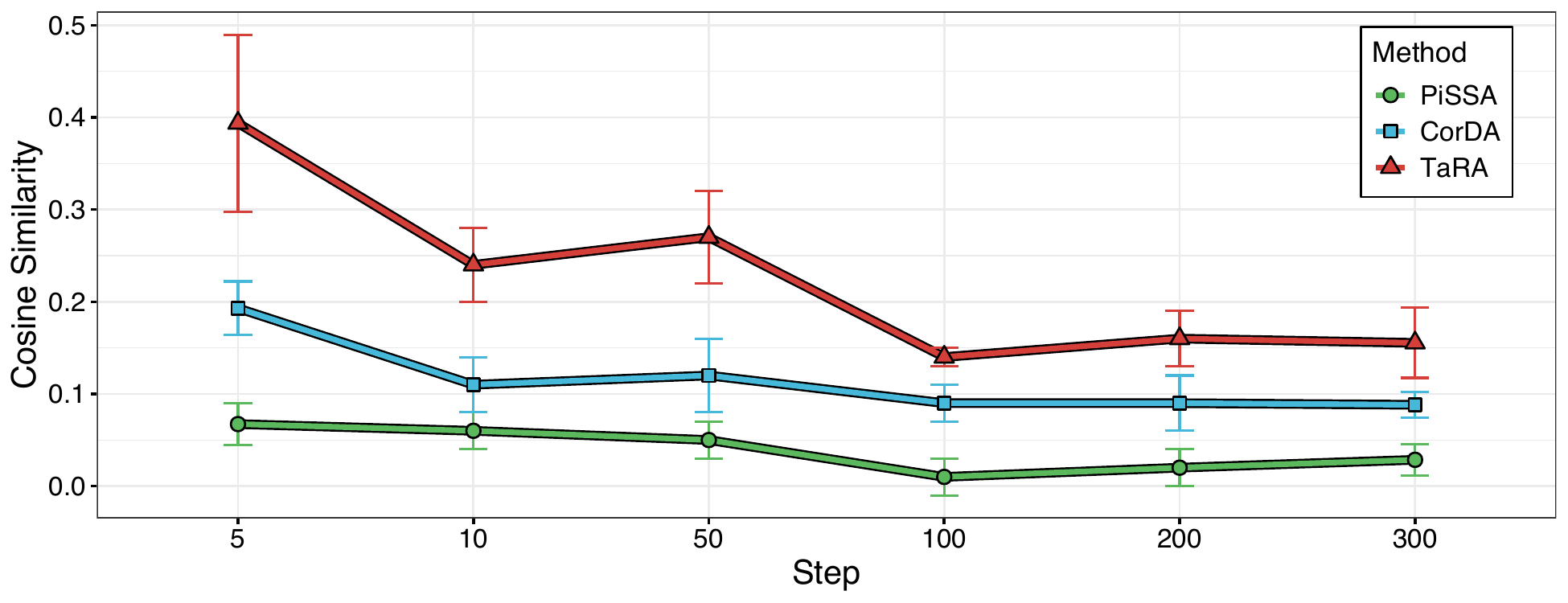}
  \caption{Gradient similarity, measured as the cosine similarity between the n-step (5, 10, 50, 100, 200 and 300) gradient of the combined LoRA adapter and that of the full weight, correlates with downstream accuracy.}
    \label{fig:n_step_alignment}
  \end{figure}

To verify whether the proposed method can effectively extract principal components that are \emph{important from a training perspective}, we use the one-step gradient computed from the full-rank weight $\mathbf{W}_{0}$ as a reference. We apply PiSSA, CorDA, and \ours~to obtain rank-$r$ approximations, restore them to the original parameter shape, and compute the corresponding one-step gradients.
We then measure the cosine similarity between the gradient induced by each low-rank parameterization and the gradient of the corresponding full-rank weight to quantify their local gradient alignment.

We conduct this experiment on RoBERTa-base \cite{liu2019roberta} trained for one step on the CoLA task \cite{warstadt2019cola} in the GLUE benchmark \cite{wang2018glue}. Cosine similarities are computed for all linear layers over 100 random samples and averaged across layers.

As shown in Figure~\ref{fig:motivation}, \ours~achieves substantially higher alignment with the full-rank gradient than PiSSA and CorDA across all ranks. PiSSA shows little improvement as the rank increases, while CorDA exhibits only modest gains due to its task-aware initialization. In contrast, \ours~demonstrates a sharp increase in cosine similarity with rank and consistently attains the highest alignment, resulting in better fine-tuning quality. These results indicate that \ours~more effectively preserves training-relevant  principal components that are critical for optimization.

\edit{Additionally, we further analyzed the effect of initialization beyond the very early stage of training. Figure \ref{fig:n_step_alignment} presents the gradient alignment measured after
n training steps (5, 10, 50, 100, 200, and 300) under the same setup as the one-step analysis above. We observe that \ours~maintains substantially higher alignment than the other methods even after multiple training steps. We also found an interesting trend in this experiment: although the gradient similarity decreases from its initial level as training progresses, after a certain point it no longer drops and instead maintains a consistent level of similarity to Full FT\footnote{We provide an intuitive explanation of this phenomenon in Appendix~\ref{app:gradient_persistence}.}. These results imply that, if the initialization induces gradients similar to those of Full FT at the beginning of training, its effect may persist throughout the training process.}

\begin{table}[t]
  \centering
  \small
  \setlength{\tabcolsep}{4pt}
  \renewcommand{\arraystretch}{1.08}
  \begin{tabular}{@{}lccc@{}}
    \toprule
    \multicolumn{1}{l}{$(r=128)$} & \textbf{Init Time} & \multicolumn{2}{c}{\textbf{Training Time}} \\
    \cmidrule(lr){2-2}\cmidrule(lr){3-4}
    & & \textbf{LoRA} & \textbf{\ours} \\
    \cmidrule(lr){3-4}
    MetaMathQA     & 18m & 6h01m & 6h20m (5\%$\uparrow$) \\
    CodeFeedback  & 18m & 6h40m & 6h58m (4\%$\uparrow$) \\
    \bottomrule
  \end{tabular}
  \caption{Initialization and training time comparison.}
  \label{tab:init_time_analysis}
\end{table}

\begin{figure}[t]
  \centering
  \begin{subfigure}[t]{0.48\columnwidth}
      \centering
      \includegraphics[width=\linewidth]{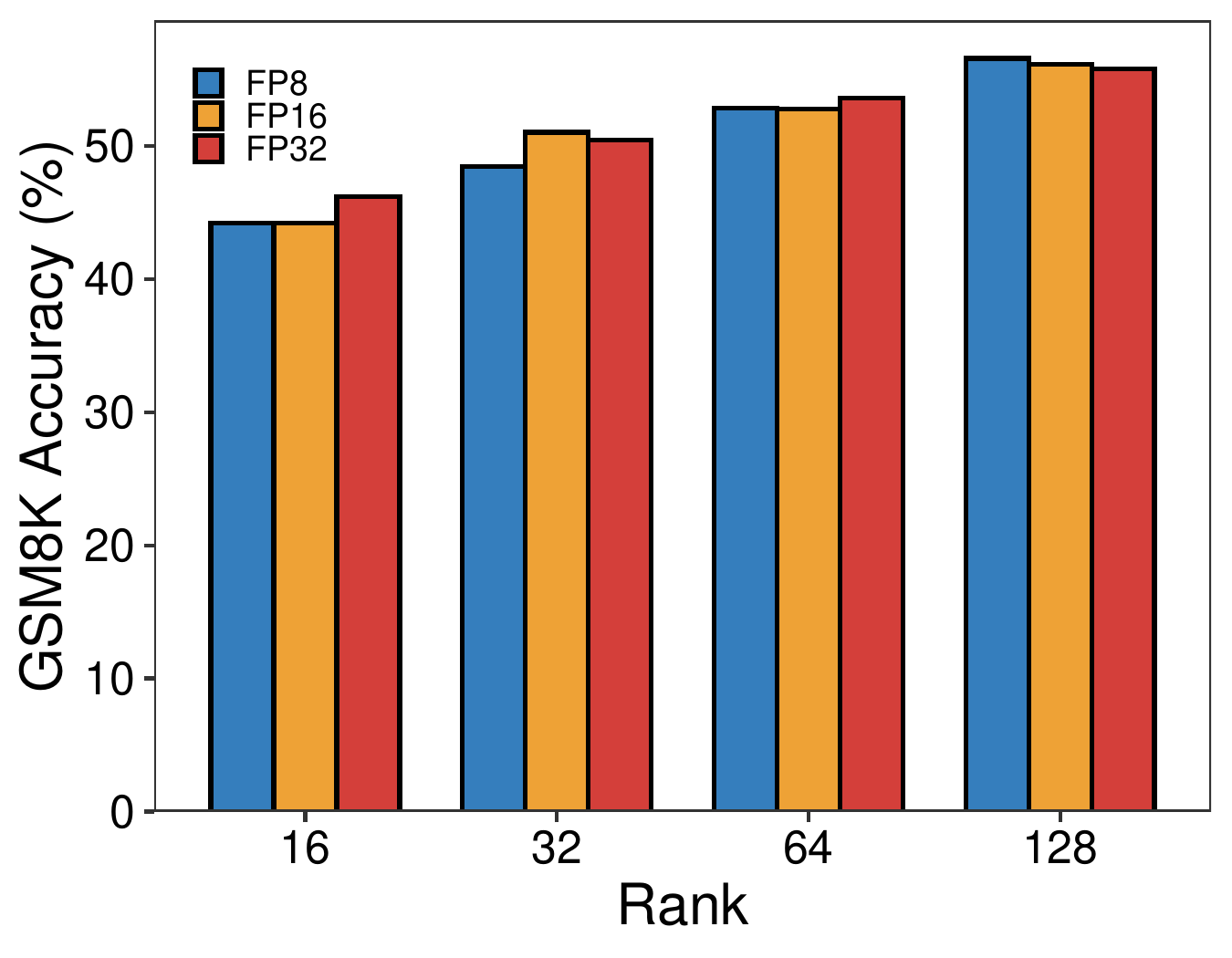}
      \caption{init precision ablation}
      \label{fig:init_precision_ablation}
  \end{subfigure}
\begin{subfigure}[t]{0.48\columnwidth}
  \centering
  \includegraphics[width=\linewidth]{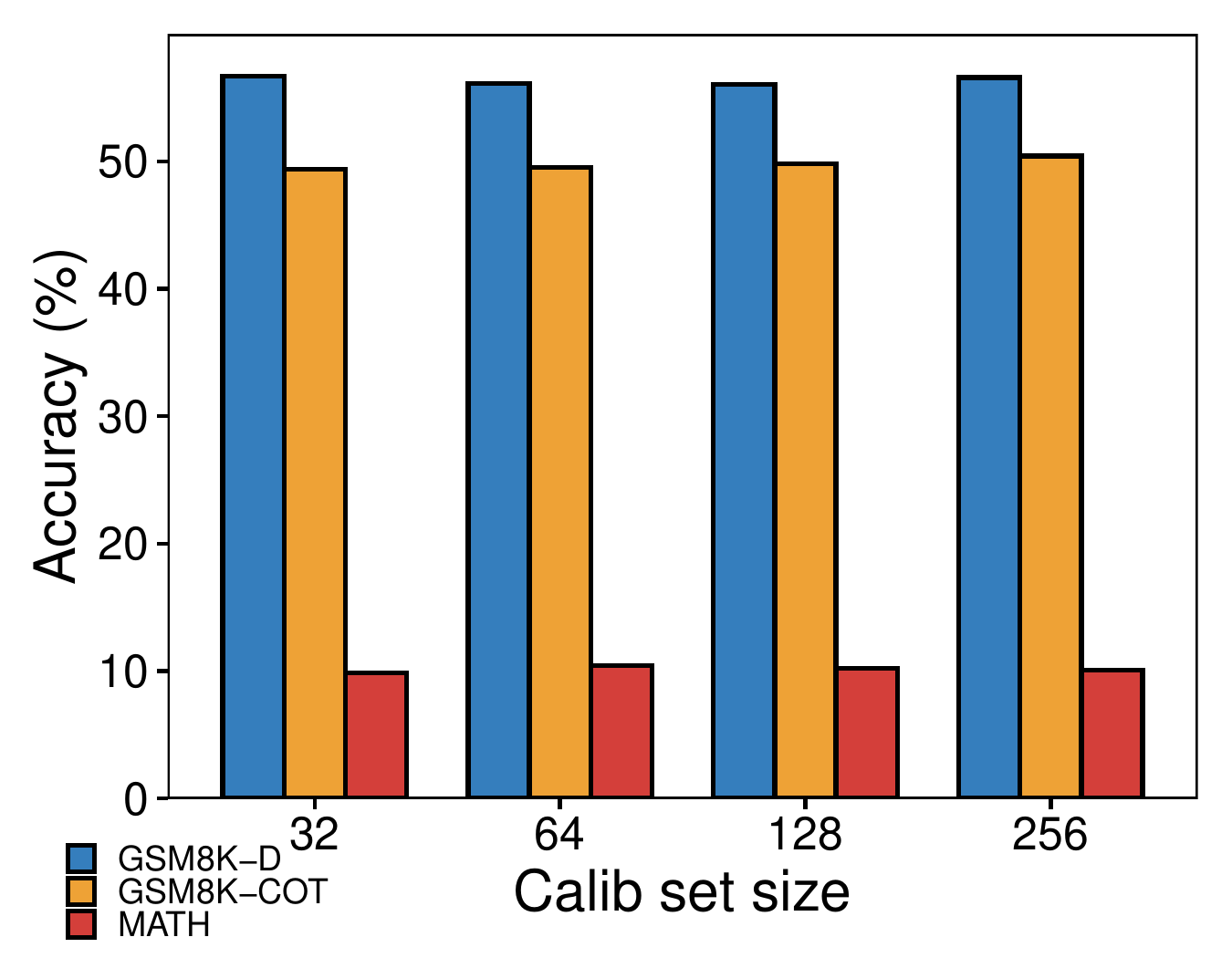}
  \caption{calib size ablation}
  \label{fig:calib_size_ablation}
\end{subfigure}
\caption{(a) Ablation of the numeric precision used to collect covariance statistics for \ours~initialization (FP8/FP16/FP32), reporting GSM8K-D accuracy across ranks $r\in{16,32,64,128}$. (b) Ablation of the calibration set size (32, 64, 128, 256), reporting math reasoning tasks accuracy.}
  \label{fig:init_precision_ablation_and_calib_size_ablation}
\end{figure}

\subsection{Initialization Overhead Analysis}

Table~\ref{tab:init_time_analysis} reports the initialization time, including covariance collection and SVD computation. For LLaMA-2-7B on MetaMathQA and CodeFeedback, initialization accounts for only 4-5\% of the total fine-tuning time, demonstrating that its computational overhead is affordable in practice.

Since \ours~requires collecting both activation and gradient covariances, its initialization can incur nontrivial memory overhead. To reduce this cost, we explore lower-precision accumulation of covariance statistics. Figure~\ref{fig:init_precision_ablation} shows the results in which covariances are collected in FP8, FP16, or FP32, followed by initialization and evaluation on LLaMA-2-7B for GSM8K-D across LoRA ranks. Lowering the precision results in only minor accuracy variations, with slight gains or losses depending on the rank. These results indicate that low-precision covariance collection is an effective alternative to reduce memory usage during initialization.

\edit{Additionally, we analyzed the performance variation with respect to the calibration set size. Figure \ref{fig:calib_size_ablation} shows the performance trends across different calibration set sizes (32,64,128, and 256). The results show that \ours{} achieves stable performance even with a small calibration set, with only minor variation in performance. This observation suggests that \ours{} remains effective even when only a small calibration set is available.}

\section{Related Work}

A wide range of PEFT methods have been proposed to reduce the cost of adapting large language models \cite{ding2023peft1,xu2023peft2}. These methods span several paradigms, including adapter-based approaches \cite{houlsby2019parameter,lei2023adapter1,he2021adapter2,ruckle2021adapter3,zhao2022adpater4,pfeiffer2021adapter5}, prompt-based methods \cite{li2021prefix,hambardzumyan2021prompt1,lester2021prompt2,vu2022prompt3,asai2022prompt4}, and low-rank adaptation techniques \cite{hu2022lora,zhang2023adalora,valipour2023dylora}. Adapter- and prompt-based methods introduce additional trainable components and optimize only these parameters, but they modify the model architecture and often incur inference overhead.

LoRA \cite{hu2022lora} addresses this limitation by freezing pretrained weights and learning a low-rank update parameterized by matrices $\mathbf{A}$ and $\mathbf{B}$, which project activations into and out of a bottleneck space. LoRA is simple, computationally efficient, and allows the learned update to be merged into the base weights after training, enabling inference without additional overhead.

Building on LoRA, many extensions have been developed to further improve efficiency and performance. These efforts primarily focus on allocating update capacity across layers and enabling training under resource constraints, particularly memory limitations. Representative approaches include adaptive rank allocation \cite{zhang2023adalora,valipour2023dylora}, redesigning low-rank update structures \cite{liu2024dora,zhao2024galore,qiu2023controlling}, and integrating LoRA with pruning \cite{zhang2024loraprune} or quantization \cite{dettmers2023qlora,xu2023qalora}. Together, these methods mitigate the accuracy degradation caused by rank constraints while preserving LoRA's core benefit.

Notably, these advances are largely orthogonal to the choice of LoRA initialization. As a result, they are complementary to the method proposed in this work, and combining them may yield further performance gains.

\section{Conclusion}
In this paper, we propose training-aware low-rank adaptation initialization (TaRA), a new PEFT approach that explicitly prioritizes training-relevant directions. TaRA performs a covariance-aware SVD that jointly incorporates activation and gradient statistics to identify directions that are most influential during optimization. Using the resulting low-rank approximation, we show that the induced one-step gradient closely matches that of full fine-tuning, indicating effective preservation of training-relevant information. When applied to LoRA initialization, TaRA consistently outperforms existing methods and achieves state-of-the-art performance across a wide range of benchmarks.

\section{Limitations}

First, \ours~is tailored to the calibration and training distribution, so its gains can be less stable under distribution shift. HumanEval and MBPP are out-of-distribution relative to the CodeFeedback training data, and their performance varies non-monotonically with rank across the compared methods (Appendix~\ref{mitigating-ood-task-acc-fluctuation}). Although \ours~remains effective on in-distribution tasks, improving its robustness to unseen distributions is an important direction for future work.

Second, compared with calibration-free methods such as LoRA and PiSSA, \ours~requires a task-specific calibration stage to collect activation and gradient covariance statistics. This introduces additional data access and one-time computation and memory costs before fine-tuning. Our overhead and calibration size analyses show that these costs are manageable in practice, but reducing this calibration dependency would further improve the applicability of \ours.

\section*{Acknowledgements}

This work was supported by IITP and NRF grant funded by the Korea government(MSIT) (RS-2024-00415602, RS-2023-00228970, RS-2019-II191906).

\bibliography{references}

\clearpage

\appendix

\onecolumn
\section*{Appendix}

\section{Proof of Equation}
\label{sec:appendix_proof_of_equations}

\paragraph{Second-order Taylor expansion.}
Let $\mathcal{L}$ be the loss and $\theta$ the parameter matrix of a model, with reference point $\theta_0$.
A second-order Taylor expansion of $\mathcal{L}(\theta)$ around $\theta_0$ gives
\begin{equation*}
\mathcal{L}(\theta)
\approx
\mathcal{L}(\theta_0)
+ \nabla\mathcal{L}(\theta_0)^{\top}(\theta - \theta_0)
+ \frac{1}{2}
(\theta - \theta_0)^{\top}
H
(\theta - \theta_0),
\qquad
H := \nabla^2 \mathcal{L}(\theta_0).
\end{equation*}

\paragraph{Hessian to Fisher curvature surrogate.}
Instead of explicitly forming the Hessian $H$, we use the Fisher information matrix $\mathcal{F}$
as a curvature surrogate, and replace the quadratic curvature term by $\mathcal{F}$:
\begin{equation*}
\mathcal{L}(\theta)
\approx
\mathcal{L}(\theta_0)
+ \nabla\mathcal{L}(\theta_0)^{\top}(\theta - \theta_0)
+ \frac{1}{2}
(\theta - \theta_0)^{\top}
\mathcal{F}
(\theta - \theta_0).
\end{equation*}

\paragraph{Fisher to K-FAC approximation.}
Following the K-FAC approximation \citep{martens2015optimizing}, we factorize the Fisher matrix as
\begin{equation*}
\mathcal{F} \approx \Sigma_X \otimes \Sigma_G,
\end{equation*}
where $\Sigma_X$ and $\Sigma_G$ are the activation and gradient covariance matrices, respectively.
Using standard Kronecker/vectorization identities, the quadratic form induced by $\mathcal{F}$ can be written as
\begin{equation*}
(\theta - \theta_0)^{\top}(\Sigma_X \otimes \Sigma_G)(\theta - \theta_0)
=
\mathrm{tr}\!\left(
(\theta - \theta_0)^{\top}\Sigma_G(\theta - \theta_0)\Sigma_X
\right).
\end{equation*}

\paragraph{Quadratic term.}
We define the corresponding curvature-induced quadratic term as
\begin{equation*}
P(\theta)
:=
\frac{1}{2}\,
\mathrm{tr}\!\left(
\Sigma_G(\theta-\theta_0)\Sigma_X(\theta-\theta_0)^{\top}
\right).
\end{equation*}
Then the Taylor model can be rewritten as
\begin{equation*}
\mathcal{L}(\theta)
\approx
\mathcal{L}(\theta_0)
+ \nabla\mathcal{L}(\theta_0)^{\top}(\theta - \theta_0)
+ P(\theta).
\end{equation*}

\paragraph{Differential/trace computation of $\nabla P(\theta)$.}
Let $E := \theta-\theta_0$. Then
\begin{equation*}
P(\theta)
=
\frac{1}{2}\,
\mathrm{tr}\!\left(
\Sigma_G E \Sigma_X E^{\top}
\right).
\end{equation*}
Taking the differential yields
\begin{equation*}
dP(\theta)
=
\frac{1}{2}\mathrm{tr}\!\left(\Sigma_G\, dE\, \Sigma_X\, E^\top\right)
+
\frac{1}{2}\mathrm{tr}\!\left(\Sigma_G\, E\, \Sigma_X\, dE^\top\right).
\end{equation*}
Using cyclicity of trace and $\mathrm{tr}(A\,dE^\top)=\mathrm{tr}(A^\top dE)$, we obtain
\begin{equation*}
dP(\theta)
=
\frac{1}{2}\mathrm{tr}\!\left(\Sigma_X E^\top \Sigma_G\, dE\right)
+
\frac{1}{2}\mathrm{tr}\!\left((\Sigma_G E \Sigma_X)^\top dE\right).
\end{equation*}
Since $\Sigma_X$ and $\Sigma_G$ are covariance matrices, we use symmetric estimates so that
$\Sigma_X^\top=\Sigma_X$ and $\Sigma_G^\top=\Sigma_G$, which makes the two terms equal and yields
\begin{equation*}
dP(\theta)
=
\mathrm{tr}\!\left((\Sigma_G E \Sigma_X)^\top dE\right).
\end{equation*}

\paragraph{Identifying the gradient and the final relation.}
By the defining relation for a scalar function,
\begin{equation*}
dP(\theta) = \mathrm{tr}\!\left((\nabla_E P(\theta))^\top dE\right),
\end{equation*}
we identify
\begin{equation*}
\nabla_E P(\theta) = \Sigma_G E \Sigma_X.
\end{equation*}
Finally, since $E=\theta-\theta_0$ implies $dE=d\theta$, we conclude
\begin{equation*}
\nabla P(\theta)
=
\Sigma_G(\theta-\theta_0)\Sigma_X.
\end{equation*}
Taking the gradient of the Taylor model
$\mathcal{L}(\theta)\approx \mathcal{L}(\theta_0)+\nabla \mathcal{L}(\theta_0)^\top(\theta-\theta_0)+P(\theta)$
gives
\begin{equation*}
\nabla \mathcal{L}(\theta)
\approx
\nabla \mathcal{L}(\theta_0)
+
\nabla P(\theta),
\end{equation*}
and therefore
\begin{equation*}
\nabla \mathcal{L}(\theta) - \nabla \mathcal{L}(\theta_0)
\approx
\Sigma_G(\theta-\theta_0)\Sigma_X.
\end{equation*}

\twocolumn

\begin{figure}[t]
\centering
  \includegraphics[width=0.9\columnwidth]{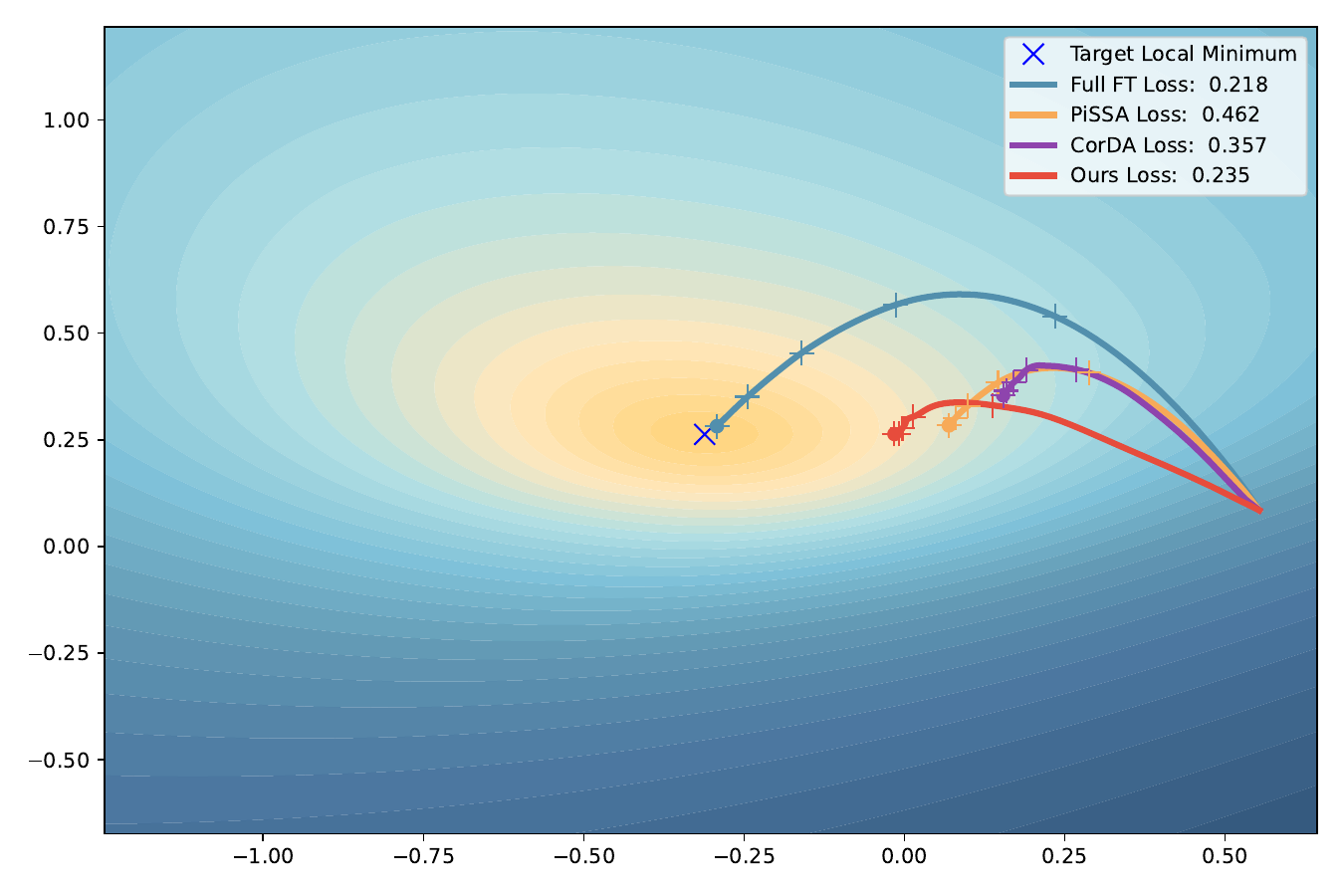}
  \caption{Loss landscape}
  \label{fig:loss_landscape}
\end{figure}

\section{GLUE Benchmark Evaluation}
\label{sec:appendix_glue_evaluation}

This section evaluates \ours~on encoder-only language models using the GLUE benchmark.

\begin{table}[t]
  \centering
  \setlength{\tabcolsep}{2pt}
  \renewcommand{\arraystretch}{1.12}
  \footnotesize
  \begin{tabular}{@{}lcccccc@{}}
  \toprule
  \textbf{Method} &
  \textbf{CoLA} &
  \textbf{SST-2} &
  \textbf{MRPC} &
  \textbf{STS-B} &
  \textbf{QNLI} &
  \textbf{AVG} \\
  \midrule
  Full FT
  & 58.67 & 93.97 & 89.61 & 89.80 & 92.23 & 84.86 \\
  \midrule
  LoRA
  & 54.68 & 94.31 & 87.25 & 88.73 & 92.52 & 83.50 \\
  PiSSA
  & 58.46 & 94.22 & 88.17 & 89.64 & 92.55 & 84.61 \\
  CorDA
  & 58.68 & 92.82 & \red{\textbf{89.07}} & 89.49 & 92.13 & 84.44 \\
  \textbf{\ours}
  & \red{\textbf{58.86}} & \red{\textbf{94.50}} & 87.50 & \red{\textbf{90.38}} & \red{\textbf{92.93}} & \red{\textbf{84.83}} \\
  \bottomrule
  \end{tabular}
  \caption{Performance on the GLUE benchmark, averaged over five seeds. \red{Red} color indicates the best PEFT result in each column, excluding Full FT.}
  \label{tab:glue}
\end{table}

We fine-tune RoBERTa-base \cite{liu2019roberta} on five GLUE tasks \cite{wang2018glue}—CoLA, SST-2, MRPC, STS-B, and QNLI—and compare Full Fine-Tuning, LoRA, PiSSA, CorDA, and \ours. Table~\ref{tab:glue} reports the mean performance over five random seeds. \ours~achieves the best results on CoLA, SST-2, STS-B, and QNLI, and also yields the strongest average performance across the five tasks. These results indicate that \ours~remains effective for encoder-only language models, not only decoder-centric LLMs.

For all PEFT methods, we use LoRA with $r=\alpha=128$, and train for 3 epochs with batch size 32.

\section{Loss-Landscape Analysis}
\label{sec:appendix_loss_landscape_analysis}

Figure~\ref{fig:loss_landscape} visualizes the gradient trajectories of Full Fine-Tuning, PiSSA, CorDA, and \ours~in a toy setting. Full Fine-Tuning and PiSSA follow a noticeably curved trajectory, taking a relatively indirect route before converging toward a local minimum. CorDA exhibits a less circuitous early trajectory, suggesting improved initial alignment. In contrast, \ours~is initialized with a direction that heads more directly toward the local minimum and ultimately converges to a lower objective value than PiSSA and CorDA. This behavior is consistent with our hypothesis: by better preserving training-relevant components, \ours~facilitates faster and more effective optimization.

This toy experiment follows the setup of PiSSA \cite{meng2024pissa}. We pre-train a three-linear-layer network on 10,000 MNIST samples from the odd-number classes, and then fine-tune on 2,000 samples from the even-number classes. We use LoRA with rank and scaling set to $r=\alpha=16$, and adopt a learning rate of $5\times10^{-4}$.

\begin{figure}[t]
  \centering
    \includegraphics[width=0.9\columnwidth]{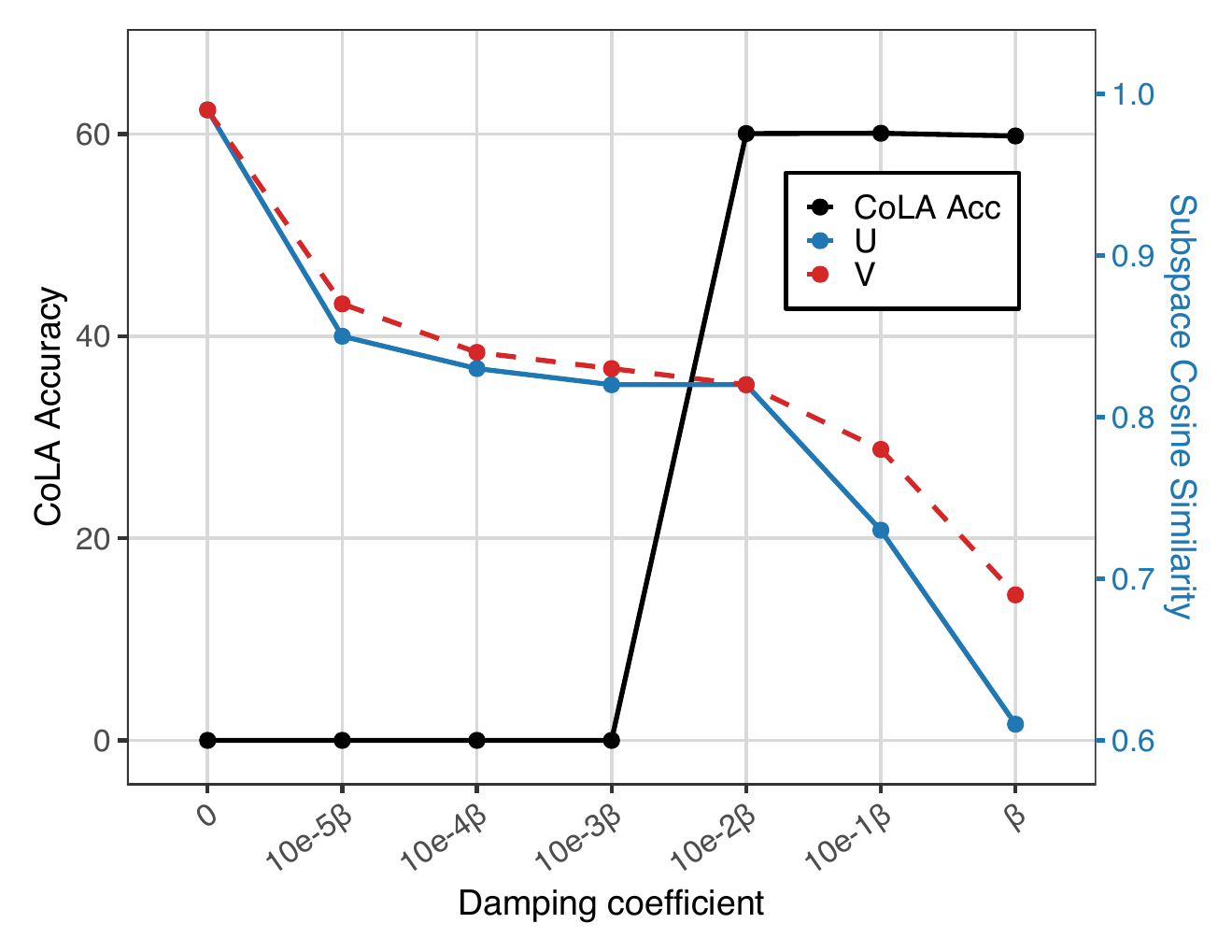}
  \caption{Damping coefficient ablation}
    \label{fig:damping_ablation}
  \end{figure}

\section{Diagonal Damping Analysis} \label{sec:diagonal_damping_analysis}

\edit{Figure \ref{fig:damping_ablation} shows the results of our CoLA-task analysis on the effect of the diagonal damping coefficient $c$, including performance stabilization due to damping (left y-axis) and the change in the resulting SVD $U$ and $V$ spaces compared with the no-damping case (right y-axis). The results show that when $c$ is smaller than $10^{-2}$, including $c=0$ (= no damping), task accuracy collapses to 0.0, indicating a failure to stabilize training and complete training breakdown. In contrast, when $c \geq 10^{-2}$, the performance becomes stable and training proceeds normally. Since $10^{-2}$ stabilizes the second-order information while preserving the original space in the no-damping case as much as possible, we choose $10^{-2}$ as an appropriate damping coefficient.}

\section{Mitigating Accuracy Fluctuations on Out-of-Distribution Tasks}
\label{mitigating-ood-task-acc-fluctuation}

\begin{table}[t]
\centering
\setlength{\tabcolsep}{3pt}
\renewcommand{\arraystretch}{1.12}
\small
\begin{tabular}{@{}clcc@{}}
\toprule
\textbf{Rank} & \textbf{Method} & \textbf{HumanEval} & \textbf{MBPP} \\
\midrule
\multirow{2}{*}{128}
& \textbf{\ours} & 22.59\std{0.58} & 25.20\std{0.67} \\
& \textbf{\ours}~+ LW ($\lambda=0.3$) & \red{\textbf{26.42}}\std{0.88} & \red{\textbf{25.67}}\std{0.38} \\
\midrule
\multirow{2}{*}{64}
& \textbf{\ours} & 23.58\std{0.70} & \red{\textbf{25.40}}\std{0.36} \\
& \textbf{\ours}~+ LW ($\lambda=0.3$) & \red{\textbf{23.98}}\std{0.75} & 25.27\std{0.90} \\
\midrule
\multirow{2}{*}{32}
& \textbf{\ours} & 21.53\std{0.29} & 24.73\std{0.52} \\
& \textbf{\ours}~+ LW ($\lambda=0.3$) & \red{\textbf{22.15}}\std{0.76} & \red{\textbf{24.83}}\std{0.19} \\
\bottomrule
\end{tabular}
\caption{Effect of Ledoit--Wolf (LW) shrinkage on code generation (OOD tasks) performance. LW results use $\lambda=0.3$. Each result is the mean over three seeds, and the standard deviation is shown in small gray text. \red{Red} color indicates the best result within each rank block.}
\label{tab:lw_shrinkage_code_results}
\end{table}

As shown in Table~\ref{tab:natural_language_generation_task_results}, the accuracies on HumanEval and MBPP do not exhibit a consistent trend as the LoRA rank increases. Unlike the mathematical reasoning tasks, these code generation benchmarks represent an out-of-distribution (OOD) setting, where the distributions of the training and evaluation datasets differ. Since TaRA estimates activation and gradient statistics from the training data, its initialization is explicitly tailored to the training distribution. While this can facilitate effective adaptation on the training data, the resulting initialization may not necessarily be optimal under a distribution shift at evaluation time.

Ledoit--Wolf (LW) shrinkage~\cite{ledoit2003lwshrinkage} has been widely used to stabilize covariance estimation when only a limited number of samples are available. In our setting, we apply shrinkage in the following form:
$$
\widetilde{\Sigma}
=
(1-\lambda)\widehat{\Sigma}_{\mathrm{calib}}
+
\lambda I,
$$
\noindent
where $I$ denotes the identity matrix. The same shrinkage can be directly incorporated into the covariance estimation procedure of \ours. The purpose of this experiment is to reduce the extent to which \ours's training-aware initialization is biased toward the calibration distribution. As $\lambda$ increases, the estimated covariance becomes progressively less dependent on the calibration data. In the limit as $\lambda$ approaches 1, the covariance becomes data-agnostic, and the resulting initialization coincides with PiSSA under our formulation.

Table~\ref{tab:lw_shrinkage_code_results} reports the code generation results obtained by estimating the covariance matrices with LW shrinkage, initializing the LoRA adapters using \ours, and subsequently training them under the same setting. Compared with the results without shrinkage, incorporating LW shrinkage yields substantially more consistent and nearly monotonic performance across different ranks. These results suggest that regularizing the training-derived covariance statistics can improve the robustness of TaRA under distribution shift, particularly for OOD code generation tasks.

\section{Why One-Step Gradient Alignment Persists Beyond Initialization}
\label{app:gradient_persistence}

Although TaRA is derived from a one-step gradient matching objective,
Figure~\ref{fig:n_step_alignment} shows that its gradient
alignment with Full FT remains substantially higher than those of the
baselines even after multiple optimization steps.
As a possible intuition, we consider an idealized setting in which
Full FT and TaRA follow the same local training dynamics. We use the
local quadratic approximation adopted in
Section~\ref{sec:training-relevant-decomposition}.

Let

$$
g_t := \nabla L(\theta_t),
$$

and consider gradient descent with learning rate $\eta$:

$$
\theta_{t+1} = \theta_t - \eta g_t.
$$

Under the local quadratic approximation around $\theta_0$ in Eq.~\ref{eq:taylor_expansion},
the gradient can be approximated as

$$
\nabla L(\theta)
\approx
\nabla L(\theta_0)
+
H(\theta-\theta_0),
$$

where $H := \nabla^2 L(\theta_0)$ denotes the local Hessian.
Therefore,

$$
\begin{aligned}
g_{t+1}
&=
\nabla L(\theta_{t+1}) \\
&\approx
\nabla L(\theta_0)
+
H(\theta_{t+1}-\theta_0) \\
&=
\nabla L(\theta_0)
+
H(\theta_t-\theta_0)
-
\eta H g_t \\
&\approx
g_t-\eta H g_t \\
&=
(I-\eta H)g_t,
\end{aligned}
$$

\noindent
where $I$ denotes the identity matrix. Repeatedly applying this relation gives

$$
g_t
\approx
(I-\eta H)^t g_0.
$$

This relation provides a simple interpretation of the empirical
behavior in Figure~\ref{fig:n_step_alignment}. Under a locally
stable curvature, later gradients are not generated along arbitrary,
unrelated directions; instead, they are obtained by repeatedly applying
a polynomial in the local Hessian to the initial gradient. This is
consistent with prior observations that training gradients tend to
concentrate in a small subspace associated with dominant Hessian
directions, and that this subspace remains relatively stable during
training~\cite{gur2018gradient}.

Consequently, if an initialization captures training-relevant
curvature directions at the beginning of optimization, these directions
can remain relevant over subsequent steps. Since TaRA explicitly
constructs its initialization using curvature-aware activation and
gradient statistics, its high one-step gradient alignment can therefore
provide a useful initialization bias beyond the first optimization
step. This local analysis is intended as an intuitive explanation
rather than a formal guarantee for the entire nonlinear training
trajectory.

\section{Calibration Set Size and LoRA alpha Ablation}
\label{sec:appendix_calibration_set_size_analysis}

\begin{table}[t]
  \centering
  \setlength{\tabcolsep}{5pt}
  \renewcommand{\arraystretch}{1.12}
  \small
  \begin{tabular}{@{}cccc@{}}
    \toprule
    \textbf{Calib Size} & \textbf{GSM8K-D} & \textbf{GSM8K-COT} & \textbf{MATH} \\
    \midrule
    32  & \red{\textbf{56.69}}\std{0.22} & 49.38\std{0.11} & 9.88\std{0.03} \\
    64  & 56.11\std{0.33} & 49.55\std{0.22} & \red{\textbf{10.41}}\std{0.04} \\
    128 & 56.05\std{0.40} & 49.80\std{0.18} & 10.23\std{0.04} \\
    256 & 56.59\std{0.25} & \red{\textbf{50.42}}\std{0.13} & 10.08\std{0.10} \\
    \bottomrule
  \end{tabular}
  \caption{Ablation on calibration set size. Each result is the mean over three seeds, and the standard deviation is shown in small gray text. \red{Red} color indicates the best result in each column.}
  \label{tab:calib_size}
\end{table}

\edit{Table \ref{tab:calib_size} shows the performance variation across different calibration set sizes at rank 128. The results indicate that \ours~remains stable even with small calibration sets, with only minor performance changes. This suggests that \ours~can still produce meaningful results even when the calibration set must be kept small due to limited available data.}

\edit{Table \ref{tab:lora_alpha} shows the performance variation across different LoRA $\alpha$ values at rank 128. The results show that \ours~achieves the best performance when LoRA $\alpha$ is set equal to the LoRA rank.}

\begin{table}[t]
  \centering
  \setlength{\tabcolsep}{5pt}
  \renewcommand{\arraystretch}{1.12}
  \small
  \begin{tabular}{@{}cccc@{}}
    \toprule
    \textbf{LoRA $\alpha$} & \textbf{GSM8K-D} & \textbf{GSM8K-COT} & \textbf{MATH} \\
    \midrule
    64  & 54.59\std{0.16} & 47.34\std{0.14} & 9.98\std{0.00} \\
    128 & \red{\textbf{56.59}}\std{0.25} & \red{\textbf{50.42}}\std{0.13} & 10.08\std{0.10} \\
    256 & 55.04\std{0.11} & 49.53\std{0.20} & \red{\textbf{10.21}}\std{0.09} \\
    \bottomrule
  \end{tabular}
  \caption{Ablation on LoRA $\alpha$. Each result is the mean over three seeds, and the standard deviation is shown in small gray text. \red{Red} color indicates the best result in each column.}
  \label{tab:lora_alpha}
\end{table}

\section{Difference between Fisher-based Model Compression and \ours}
\label{sec:difference_between_compression_and_ours}

\begin{table}[t]
  \centering
  \setlength{\tabcolsep}{0.5pt}
  \renewcommand{\arraystretch}{1.12}
  \footnotesize
  \begin{tabular}{@{}lccc@{}}
    \toprule
    \textbf{Variant} & \textbf{GSM8K-D} & \textbf{GSM8K-COT} & \textbf{MATH} \\
    \midrule
    with sqrt & 53.12\std{0.30} & 48.28\std{0.19} & 8.97\std{0.11} \\
    without sqrt (\textbf{\ours}) & \red{\textbf{56.59}}\std{0.25} & \red{\textbf{50.42}}\std{0.13} & \red{\textbf{10.08}}\std{0.10} \\
    \bottomrule
  \end{tabular}
  \caption{Ablation on the square-root scaling term. Each result is the mean over three seeds, and the standard deviation is shown in small gray text. \red{Red} color indicates the best result in each column.}
  \label{tab:sqrt_ablation}
\end{table}

\edit{Recent LLM compression methods perform compression using Fisher information \cite{hsu2207language, hua2022numerical,chekalina2025gfwsvd}. From the compression perspective, the following formulation is used:}

\[
\min_{\theta}\,\mathcal{L}(\theta) - \mathcal{L}(\theta_0)
\approx
\min_{\theta}\,\left\| \mathcal{F}^{1/2}(\theta - \theta_0) \right\|_F^2.
\]

\edit{It uses a formulation that finds $\theta$ by minimizing the change in the loss function $\mathcal{L}$. Specifically, it seeks $\theta$ that remains as close as possible to the original model $\theta_{0}$, weighted by the square-root Fisher information matrix $\mathcal{F}^{1/2}$.}

\edit{In contrast, \ours~uses a formulation that finds $\theta$ by minimizing the change in the gradient:}

\[
\min_{\theta}\,\left\| \nabla \mathcal{L}(\theta) - \nabla \mathcal{L}(\theta_0) \right\|
\approx
\min_{\theta}\,\left\| \mathcal{F}(\theta - \theta_0) \right\|.
\]

\edit{It uses a formulation that finds $\theta$ by keeping it as close as possible to the original model $\theta_{0}$, weighted by the Fisher information matrix $\mathcal{F}$ without taking its square root.}

\edit{Table \ref{tab:sqrt_ablation} compares these two formulations for LoRA initialization on the MATH task at rank 128. The results show that applying LoRA initialization using the square-root formulation from the compression perspective leads to suboptimal performance. This supports the significance of the theoretical analysis of \ours{} from the training perspective.}

\section{Hyperparameters}
\label{sec:appendix_hyperparameters}

\begin{table*}[t]
\centering
\setlength{\tabcolsep}{5pt}
\renewcommand{\arraystretch}{1.12}
\small

\begin{tabularx}{0.9\textwidth}{@{}ll>{\raggedright\arraybackslash}X@{}}
\toprule
\textbf{Section} & \textbf{Hyperparameter} & \textbf{Value} \\
\midrule

\multirow{2}{*}{\textbf{LoRA}}
  & Rank $r$ (= Scaling $\alpha$)   & 32, 64, 128 \\
  & Dropout             & 0.0 \\
\midrule

\multirow{6}{*}{\textbf{Training}}
  & Epochs                      & 1 \\
  & Learning rate (\emph{except} LoRA-One)     & $4\times10^{-5}$ \\
  & Learning rate (for LoRA-One)     & $2\times10^{-4}$ \\
  & Batch size                  & 8 \\
  & Gradient accumulation steps & 16 \\
  & Seed & 0, 1, 2 \\
\midrule

\multirow{2}{*}{\textbf{Model}}
  & Model name          & \texttt{meta-llama/Llama-2-7b-hf} \\
  & Max sequence length & 512 \\
\midrule

\multirow{3}{*}{\textbf{Data}}
  & Dataset (for math)   & \texttt{meta-math/MetaMathQA} \\
  & Dataset (for code)   & \texttt{m-a-p/CodeFeedback-Filtered-Instruction} \\
  & Dataset split   & \texttt{train[:100000]} \\
\midrule

\multirow{4}{*}{\textbf{Optimization}}
  & Optimizer     & \texttt{adamw\_torch} \\
  & Weight decay  & 0.0 \\
  & Warmup ratio  & 0.03 \\
  & LR scheduler  & \texttt{cosine} \\
\bottomrule
\end{tabularx}
\caption{Hyperparameters used to fine-tune models on the natural language generation tasks.}
\label{tab:nlg_hparams_lora}
\end{table*}

\begin{table*}[t]
\centering
\setlength{\tabcolsep}{5pt}
\renewcommand{\arraystretch}{1.12}
\small

\begin{tabularx}{0.9\textwidth}{@{}ll>{\raggedright\arraybackslash}X@{}}
\toprule
\textbf{Section} & \textbf{Hyperparameter} & \textbf{Value} \\
\midrule

\multirow{2}{*}{\textbf{LoRA}}
  & Rank $r$ (= Scaling $\alpha$)   & 128 \\
  & Dropout             & 0.0 \\
\midrule

\multirow{6}{*}{\textbf{Training}}
  & Epochs                      & 1 \\
  & Learning rate (\emph{except} LoRA-One)     & $2\times10^{-5}$ \\
  & Learning rate (for LoRA-One)     & $2\times10^{-4}$ \\
  & Batch size                  & 8 \\
  & Gradient accumulation steps & 16 \\
  & Seed & 0, 1, 2 \\
\midrule

\multirow{5}{*}{\textbf{Model}}
  & Model name 1          & \texttt{deepseek-ai/DeepSeek-R1-Distill-Qwen-1.5B} \\
  & Model name 2          & \texttt{meta-llama/Llama-2-7b-hf} \\
  & Model name 3          & \texttt{meta-llama/Llama-3.1-8B} \\
  & Model name 4          & \texttt{Qwen/Qwen3-8B} \\
  & Max sequence length & 256 \\
\midrule

\multirow{1}{*}{\textbf{Data}}
  & Dataset   & \texttt{zwhe99/commonsense\_170k} \\
\midrule

\multirow{4}{*}{\textbf{Optimization}}
  & Optimizer     & \texttt{adamw\_torch} \\
  & Weight decay  & 0.0 \\
  & Warmup ratio  & 0.03 \\
  & LR scheduler  & \texttt{cosine} \\
\bottomrule
\end{tabularx}
\caption{Hyperparameters used to fine-tune models on the natural language understanding tasks.}
\label{tab:nlu_hparams_lora}
\end{table*}

This section documents the hyperparameters used in Section~\ref{sec:experiemnt}. Table~\ref{tab:nlg_hparams_lora} lists the hyperparameters for the natural language generation experiments, while Table~\ref{tab:nlu_hparams_lora} reports those used for the natural language understanding experiments.

\end{document}